\documentclass[final,5p,times,twocolumn]{elsarticle}
\usepackage{multirow}
\usepackage{amssymb}
\usepackage{amsmath}
\usepackage{booktabs}
\usepackage{tabularx}
\usepackage{array}
\usepackage{longtable}
\usepackage{color}
\usepackage{subfigure} 
\usepackage{algorithmic}
\usepackage{algorithm}
\usepackage{array}
\usepackage{textcomp}
\usepackage{stfloats}
\usepackage{url}
\usepackage{verbatim}
\usepackage{graphicx}
\usepackage{xcolor}
\usepackage{subcaption}
\usepackage{color}
\usepackage[marginal]{footmisc}

\begin{document}

\begin{frontmatter}

\title{TemPrompt: Multi-Task Prompt Learning for Temporal Relation Extraction in \\ RAG-based Crowdsourcing Systems }

\author[1,2]{Jing Yang\fnref{equal}}
\ead{yangjing2020@ia.ac.cn}

\author[3]{Yu Zhao\fnref{equal}}
\ead{yuzhao@nudt.edu.cn}

\author[4]{Linyao Yang}
\ead{yangly@zhejianglab.com}

\author[5]{Xiao Wang\corref{corresponding}}
\ead{xiao.wang@ahu.edu.cn}

\author[6]{Long Chen}
\ead{longchen@um.edu.mo}

\author[1,2,7]{Fei-Yue Wang}
\ead{feiyue.wang@ia.ac.cn}

\cortext[corresponding]{Corresponding author.}
\fntext[equal]{The two authors contribute equally to this work.}

\address[1]{Institute of Automation, Chinese Academy of Sciences, Beijing 100190, China}
\address[2]{School of Artificial Intelligence, University of Chinese Academy of Sciences, Beijing 100049, China}
\address[3]{National Key Laboratory of Information Systems Engineering, National University of Defense Technology, Changsha 410000, China}
\address[4]{Zhejiang Lab, Hangzhou 311121, China}
\address[5]{School of Artificial Intelligence, Anhui University, Hefei 266114, China}
\address[6]{Department of Computer and Information Science, University of Macau, Macao 999078, China}
\address[7]{ Faculty of Innovation Engineering, Macau University of Science and Technology, Macao 999078, China}

\begin{abstract}
Temporal relation extraction (TRE) aims to grasp the evolution of events or actions, and thus shape the workflow of associated tasks, so it holds promise in helping understand task requests initiated by requesters in crowdsourcing systems. However, existing methods still struggle with limited and unevenly distributed annotated data. Therefore, inspired by the abundant global knowledge stored within pre-trained language models (PLMs), we propose a multi-task prompt learning framework for TRE (TemPrompt), incorporating prompt tuning and contrastive learning to tackle these issues. To elicit more effective prompts for PLMs, we introduce a task-oriented prompt construction approach that thoroughly takes the myriad factors of TRE into consideration for automatic prompt generation. In addition, we design temporal event reasoning in the form of masked language modeling as auxiliary tasks to bolster the model's focus on events and temporal cues. The experimental results demonstrate that TemPrompt outperforms all compared baselines across the majority of metrics under both standard and few-shot settings. A case study on designing and manufacturing printed circuit boards is provided to validate its effectiveness in crowdsourcing scenarios.

\end{abstract}


\end{frontmatter}
\newcommand\blfootnote[1]{%
  \begingroup
  \renewcommand\thefootnote{}\footnote{#1}%
  \addtocounter{footnote}{-1}%
  \endgroup
}

\section{Introduction}
\begin{figure}[!h]
    \centering  
    \includegraphics[width=1\linewidth]{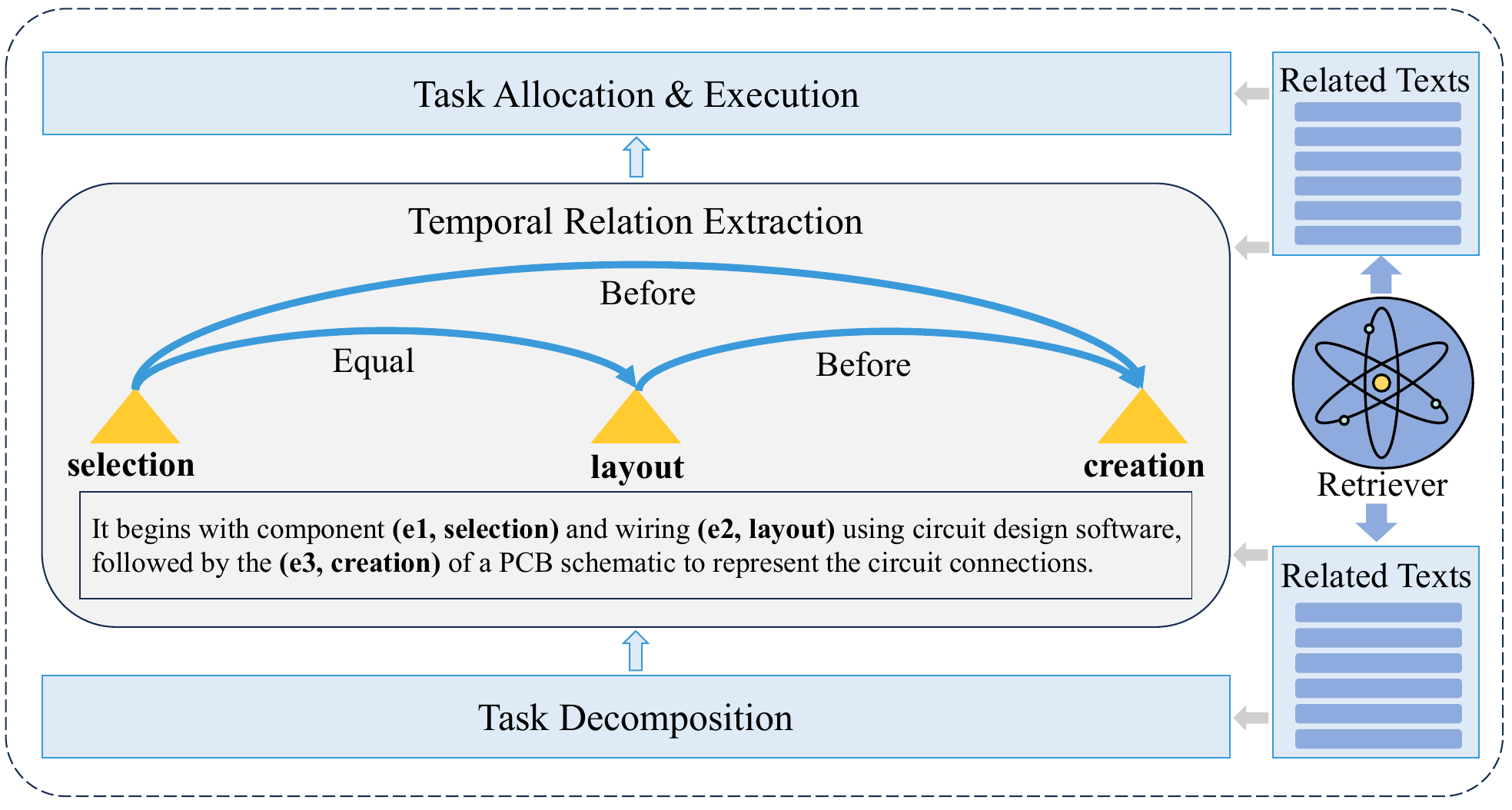}
    \caption{An example sentence and its temporal graph in RAG-based crowdsourcing systems} 
    \label{fig: graph}
\end{figure}

Temporal relation extraction (TRE) is a vital technique for comprehending event evolution and structuring task workflows, anticipated to play a pivotal role in crowdsourcing systems \cite{wang2016crowdsourcing,wang2019social,yang2024RAG} based on retrieval-augmented generation (RAG) \cite{lewis2020retrieval}. In RAG-based systems \cite{yang2024RAG}, once a task request is received, the relevant linguistic descriptions of the task execution steps will be retrieved from various databases or online search engines. By this means, the task decomposition is transformed into event detection, where each detected event is regarded as a subtask. However, the decomposed subtasks are temporally dependent, so it is necessary to extract the temporal order of events and construct their temporal graph. This is a prerequisite for the subsequent effective allocation and execution of tasks. Fig. \ref{fig: graph} illustrates an example sentence and its temporal graph in RAG-based crowdsourcing systems. There are three events in the sentence, \textit{selection}, \textit{layout} and \textit{creation}. The temporal relation between \textit{selection} and \textit{layout} is \textit{EQUAL}, and they occur \textit{BEFORE} \textit{creation}.

\begin{figure}[htbp]
\centering 
\subfigure[MATRES]{
\begin{minipage}[t]{0.45\linewidth}
\centering
\includegraphics[width=1.8in]{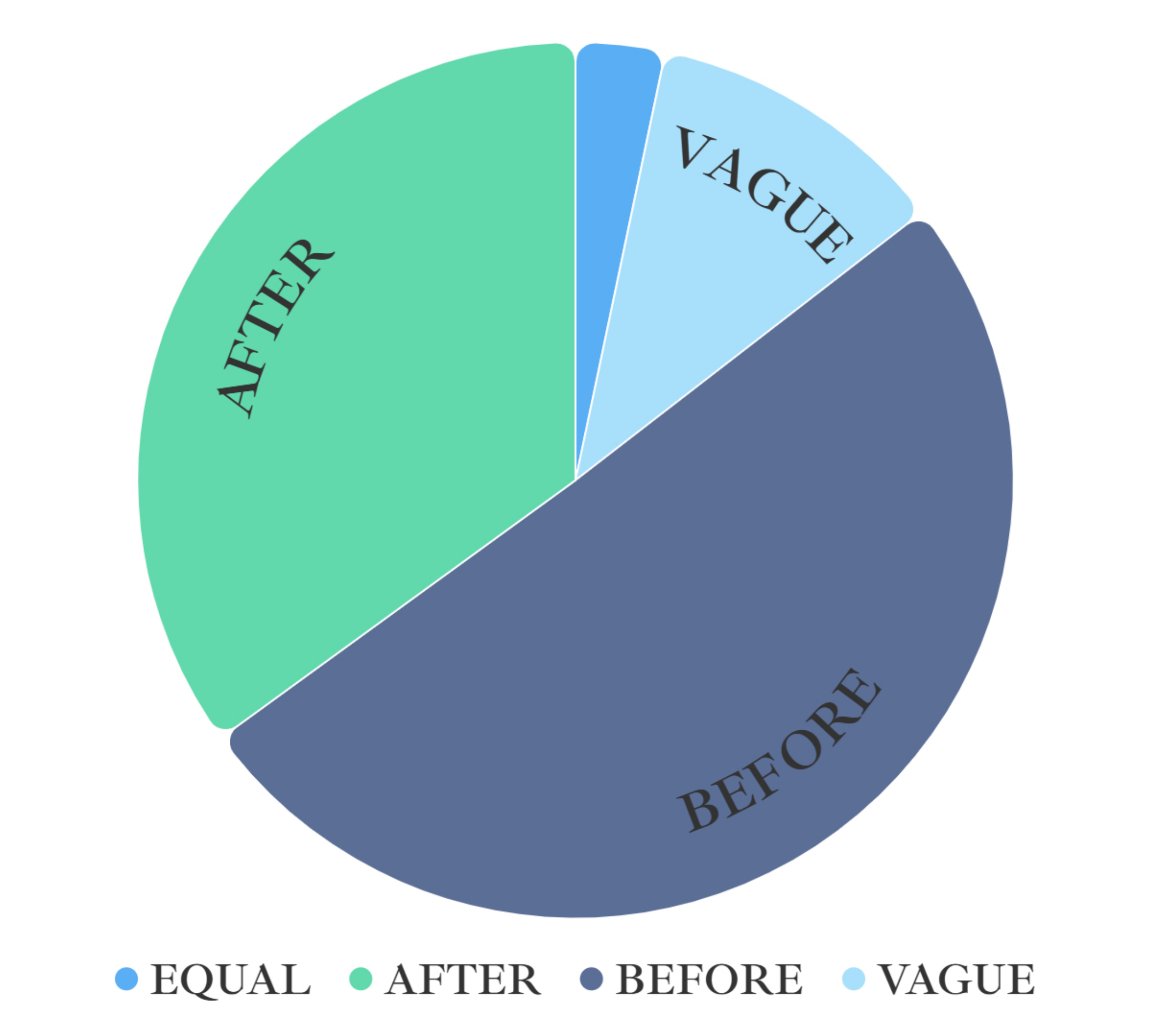}
\end{minipage}%
}%
\subfigure[TB-Dense]{
\begin{minipage}[t]{0.48\linewidth}
\centering
\includegraphics[width=1.8in]{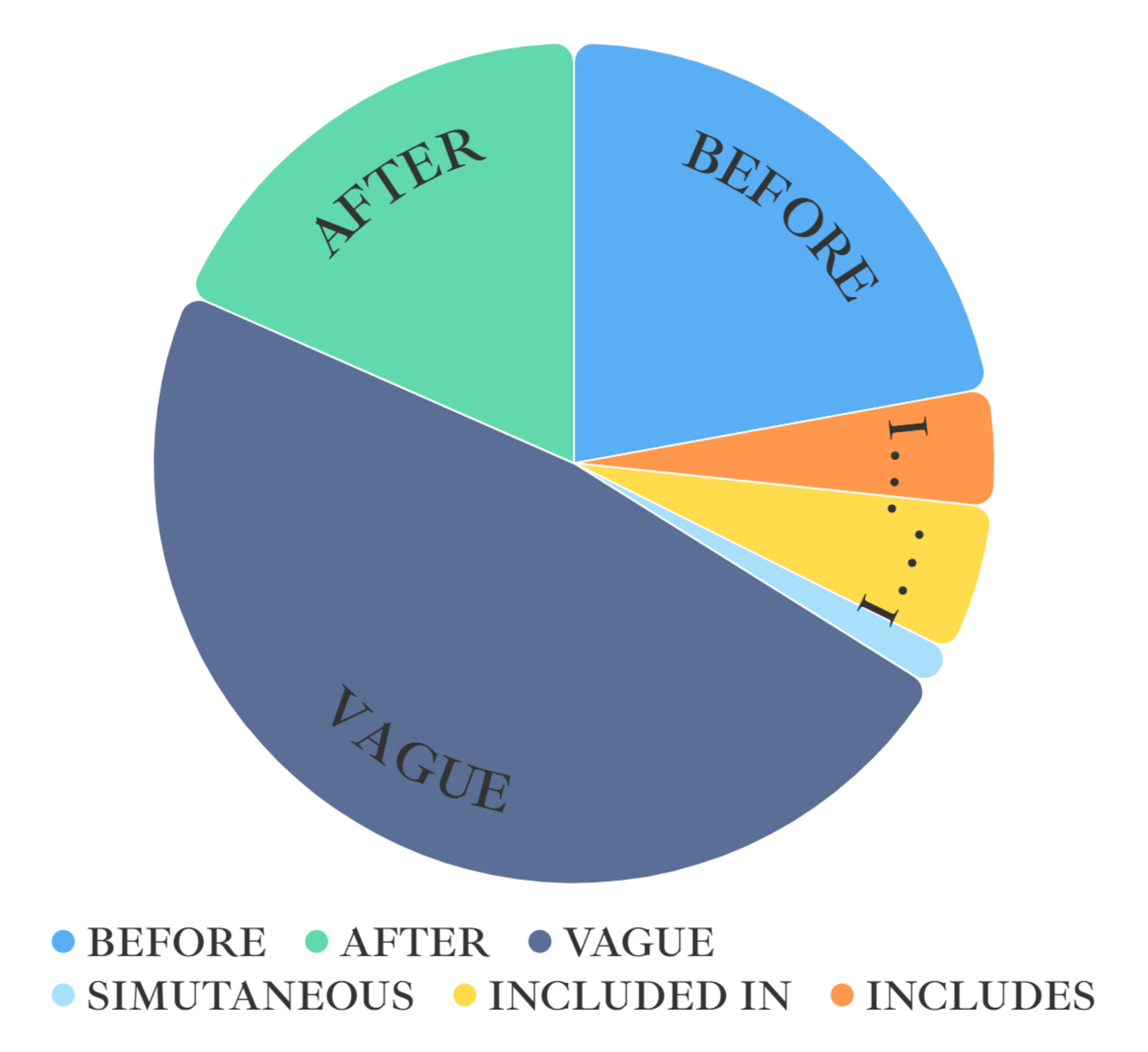}
\end{minipage}%
}%
\centering
\caption{The proportion of samples across different relation categories in the datasets.}
\label{Proportion}
\end{figure}

The existing neural network-based approaches have achieved state-of-the-art performance \cite{wen2021utilizing, cheng2017classifying, ross2020exploring,mathur2021timers,venkatachalam2021serc,zhang2021extracting,meng2017temporal}, particularly with the utilization of pre-trained language models (PLMs) like BERT \cite{devlin2018bert} for encoding. However, in real-world applications, a significant challenge faced by TRE models is the inadequacy and disparity within human-provided annotations. Fig. \ref{Proportion} demonstrates the proportion of samples across different relation categories in the MATRES\footnote{https://github.com/qiangning/MATRES} and TB-Dense\footnote{https://github.com/rujunhan/TEDataProcessing/tree/master/TBDense/timeml} dataset, two common benchmark datasets for TRE. Statistical analysis highlights a notable discrepancy in sample distribution across multiple categories. In the MATRES, \textit{BEFORE} and \textit{AFTER} vastly outnumbering \textit{VAGUE} and \textit{EQUAL}, especially, with \textit{EQUAL} occupying less than a tenth of the former two. In the TB-Dense dataset, \textit{VAGUE} accounts for nearly half of the total, while \textit{INCLUDED IN}, \textit{INCLUDED}, and \textit{SIMULTANEOUS} together comprise approximately 10\% of the total. Therefore, some efforts for data augmentation have been made \cite{ballesteros2020severing,cheng2023dynamically,zhuang2023knowledge}, such as multi-task learning, self-training, and external knowledge enhancement. However, the improvement of these methods fall short of expectations, as the introduction of additional knowledge repositories also brings in noise.

\begin{figure}[!h]
    \centering  
    \includegraphics[width=1\linewidth]{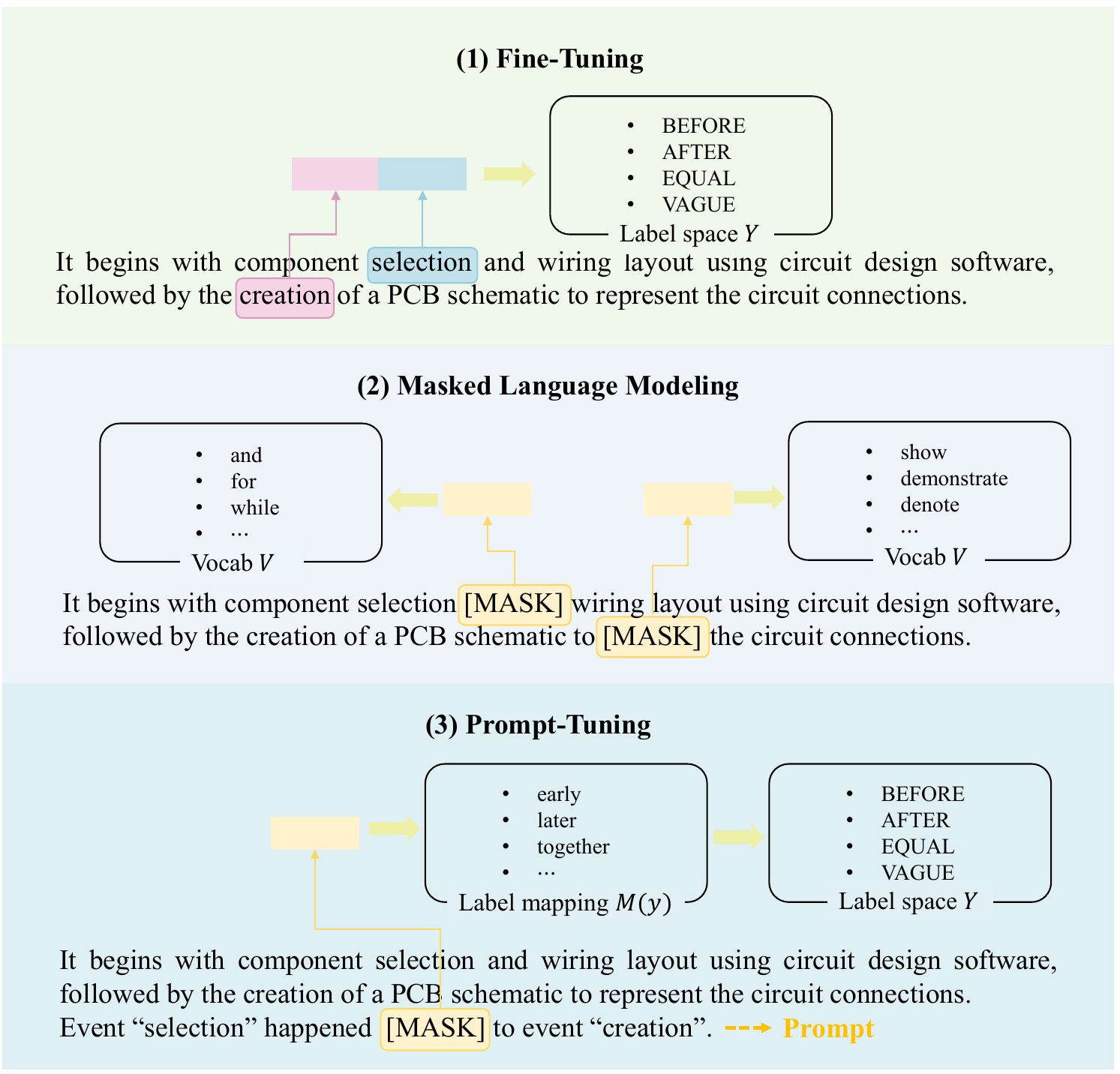}
    \caption{Comparisons of fine-tuning, masked language modeling and prompt-tuning. (1) Fine-tuning leverages the contextualized representations of two event trigger words for the direct prediction of classification labels; (2) Masked language modeling receives a partially masked sentence as input to predict the masked words and is typically employed as an auxiliary task to deepen text comprehension; (3) Prompt tuning predicts textual answers by adding a prompt containing [MASK], which are then mapped to the label space.} 
    \label{fig: tuning}
\end{figure}

PLMs have been verified to accumulate a wealth of global knowledge during the pre-training phase, promising to tackle the challenge of limited annotated data \cite{qiu2020pre}.  Fine-tuning is a common way to harness knowledge inherited within PLMs but requires an abundant supply of supervised corpora, and there are objective disparities between pre-training and it. Consequently, prompt tuning is regarded as a feasible and potential paradigm and garners widespread attention \cite{liu2023pre}. It offers hints via an additional sequence to frame the downstream tasks as masked language modeling problems, thereby bridging the gap of objective forms between pre-training and fine-tuning. Fig. \ref{fig: tuning} illustrates the main differences among fine-tuning, masked language modeling and prompt tuning. 

The key to prompt tuning is the design of cloze prompts. However, manual prompt engineering \cite{huang2023more,zhang2024event} is labor-intensive while continuous prompt learning \cite{li2021prefix,chen2022coherent}  comes with significant supervisory costs. Discrete prompt learning \cite{gao2020making,song2023taxonprompt} can automatically generate meaningful prompts under a few-shot setting, but it remains agnostic to tasks and fails to directly adapt to TRE by leveraging task-specific features such as triggers, labels and temporal attributes. Therefore, we focus on incorporating these task-specific features into the automatic design of discrete prompts for TRE and fostering a deeper understanding of temporal knowledge of events. We aim to combine prompt tuning and contrastive learning to address the inadequacy and unevenness of annotated data. Consequently, the main contributions of our work are as follows: 
\begin{itemize}
    \item We propose a multi-task prompt learning framework for TRE, named TemPrompt, which is the first study to integrate prompt tuning and contrastive learning to deal with various data issues in the field of TRE, to the best of our knowledge. 
    \item We present a task-oriented automatic generation approach for discrete prompts of TRE, which takes triggers, labels, and event mentions into consideration and requires no significant supervision costs. 
    \item We create language modeling of temporal event reasoning as auxiliary tasks during the prompt-based fine-tuning phase to enhance the model's comprehension of temporal event knowledge.
    \item We evaluate our proposed model against state-of-the-art baselines on the MATRES and TB-Dense datasets. The experimental results show that TemPrompt can achieve competitive performance.  
\end{itemize}

The following sections of this paper are organized as follows. Section \uppercase\expandafter{\romannumeral2} offers a review of the related work on TRE and prompt learning. In Section \uppercase\expandafter{\romannumeral3}, our problem statement and associated notations are introduced. Section \uppercase\expandafter{\romannumeral4} provides a thorough explanation of the model design and its components. In Section \uppercase\expandafter{\romannumeral5}, a series of experimental results are given, compared and analyzed. Section \uppercase\expandafter{\romannumeral6} draws a conclusion and offers a future outlook.

\section{Related Work}
In this section, we first give an overview of existing TRE methods. Then the current state of prompt learning, as a key technology for our work, is reviewed.  

\subsection{Temporal Relation Extraction}
Early studies primarily rely on rule/statistical-based methods \cite{chambers2014dense, bethard2007timelines,yoshikawa2009jointly,mani2006machine,fei2020cross,fei2020latent,cao2022oneee}, which utilize classical machine learning models in conjunction with handcrafted features to capture the temporal relations between events. For example, Chambers et al. \cite{chambers2014dense} introduce CAEVO, which integrates linguistic and syntactic rule-based classifiers with machine-learned classifiers using the sieve architecture. However, these methods are time-consuming, labor-intensive, and challenging to extend to other tasks or new domains. 

To deal with these issues, numerous efforts have been devoted to deep learning-based methods. The initial strategy is using various recurrent neural networks (RNNs) to capture the long and short-term dependencies within sentences \cite{meng2017temporal,ning2019improved,goyal2019embedding}. Ni et al. \cite{ning2019improved} integrate a long short-term memory (LSTM) network and a common sense encoder to improve TRE performance. Since the advent of BERT, for better contextual modeling, there has been a shift towards using PLMs like BERT to replace RNNs \cite{ross2020exploring,mathur2021timers,venkatachalam2021serc,zhang2021extracting}. Zhang et al. \cite{zhang2021extracting} leverage BERT to get a contextual presentation and thus mine temporal clues through a syntax-guided graph transformer. However, the greatest challenge in enhancing model performance is the scarcity of labeled data, particularly the imbalance in data distribution. Consequently, various methods have been proposed and attempted to address it, such as self-training \cite{cao2021uncertainty,ballesteros2020severing}, multi-task learning \cite{ballesteros2020severing,wen2021utilizing,cheng2023dynamically,zhang2022knowledge}, constructing event graphs \cite{cheng2023dynamically,liu2021discourse,vo2020extracting,lin2020conditional,fei2022matching}, external knowledge constraints and enhancement \cite{hwang2022event,han2019deep,zhou2021clinical,zhuang2023knowledge}, embedding temporal expressions \cite{goyal2019embedding,han2019joint,ma2021eventplus} as well as predicting time points \cite{huang2023more,cheng2020predicting}. Ballesteros et al. \cite{ballesteros2020severing} take relation extraction and temporal annotation datasets as complementary datasets and leverage self-training to generate a silver dataset for scheduled multitask learning. Cheng et al. \cite{cheng2023dynamically} develop a source event-centric TLINK chain and utilize all three categories (E2E, E2T and E2D) of TLINKs in the chain for multi-task learning. Han et al.\cite{han2019deep} incorporate linguistic constraints and domain knowledge through the structured support vector machine for making joint TRE predictions. Zhuang et al. \cite{zhuang2023knowledge} fuse semantic knowledge from event ontologies into natural language textual prompts to improve TRE. Goyal et al. \cite{goyal2019embedding} incorporate temporal awareness into models by embedding timexes through the use of LSTM. Huang et al. \cite{huang2023more} convert temporal relations into logical expressions of time points and thus complete the TRE by predicting the relations between certain time point pairs. These methods, however, fall short of fully leveraging the knowledge and reasoning potential of PLMs by restricting their use to context encoding alone. This limitation results in a form gap between fine-tuning and pre-training, hindering the anticipated performance gains. 

Our work investigates leveraging cloze prompts to bridge the gap between pre-training and fine-tuning, specifically targeting the PLMs’ temporal reasoning ability through the introduction of event temporal knowledge. ECONET \cite{han2020econet} is the most relevant work to ours. Still, it focuses on masking temporal indicators and related events within the original sentences as masked language models during the pre-training phase. On the other hand,  our focus is to mask events within the designed prompts during the fine-tuning phase as a temporal event reasoning task, thereby fostering the model's comprehension of temporal event knowledge expressed in the original sentence.

\subsection{Prompt Learning}
The emergence of GPT series, especially GPT-3 \cite{brown2020language}, has significantly propelled the advancement of prompt learning, exhibiting remarkable performance across various natural language processing tasks, involving relation extraction \cite{chen2022knowprompt,zhang2022prompt,zhao2024few}. The key to prompt learning lies in template construction, including manual template engineering and automated template learning. Manual template engineering is the most natural and intuitive approach, but it relies on human expertise and may face a comprehension gap between the model and humans. Huang et al. \cite{huang2023more} design Q\&A prompts to ascertain time points of events for the further prediction of temporal relations. Zhang et al. \cite{zhang2024event} leverage RAG to enhance TRE prompt templates and verbalizers based on manually crafted trigger modifiers and prompts. 

Automated template learning \cite{chen2022coherent,li2021prefix,wu2022idpg,yang2023tailor} has been proposed to overcome the challenges of manual template engineering and it can be divided into continuous and discrete prompt learning. Continuous prompt learning \cite{li2021prefix,chen2022coherent} involves generating and optimizing task-specific vectors for soft prompts in a continuous space without mapping to any concrete words. Li and Liang \cite{li2021prefix} propose prefix-tuning to optimize a sequence of “virtual tokens”. Chen et al. \cite{chen2022coherent} generate a soft prompt close to coherent texts and away from incoherent texts in the hidden space through contrastive learning for the long text generation task. However, these methods often demand costly human supervision and tend to prioritize high performance over readability, interpretability, and transferability. Moreover, some soft prompts are trained based on one or more pre-designed hard prompts. Discrete prompt learning \cite{gao2020making,song2023taxonprompt} refers to the automated exploration of templates for hard prompts within a discrete space composed of meaningful words. Gao et al. \cite{gao2020making} introduce an automated prompt generation pipeline based on seq2seq PLMs for sentence-level classification and relation prediction. Subsequently, this pipeline is applied to the generation of event classification templates by Song et al. \cite{song2023taxonprompt}. However, it is non-trivial to adapt it to TRE as TRE focuses on the temporal relations between two events corresponding to two triggers. Therefore, this paper aims to automatically generate TRE cloze prompts by maintaining the fixed order of triggers to ensure accurate label correspondence under the simultaneous consideration of triggers, labels, and sentences.


\begin{figure*}[!h]
    \centering  
    \includegraphics[width=1\linewidth]{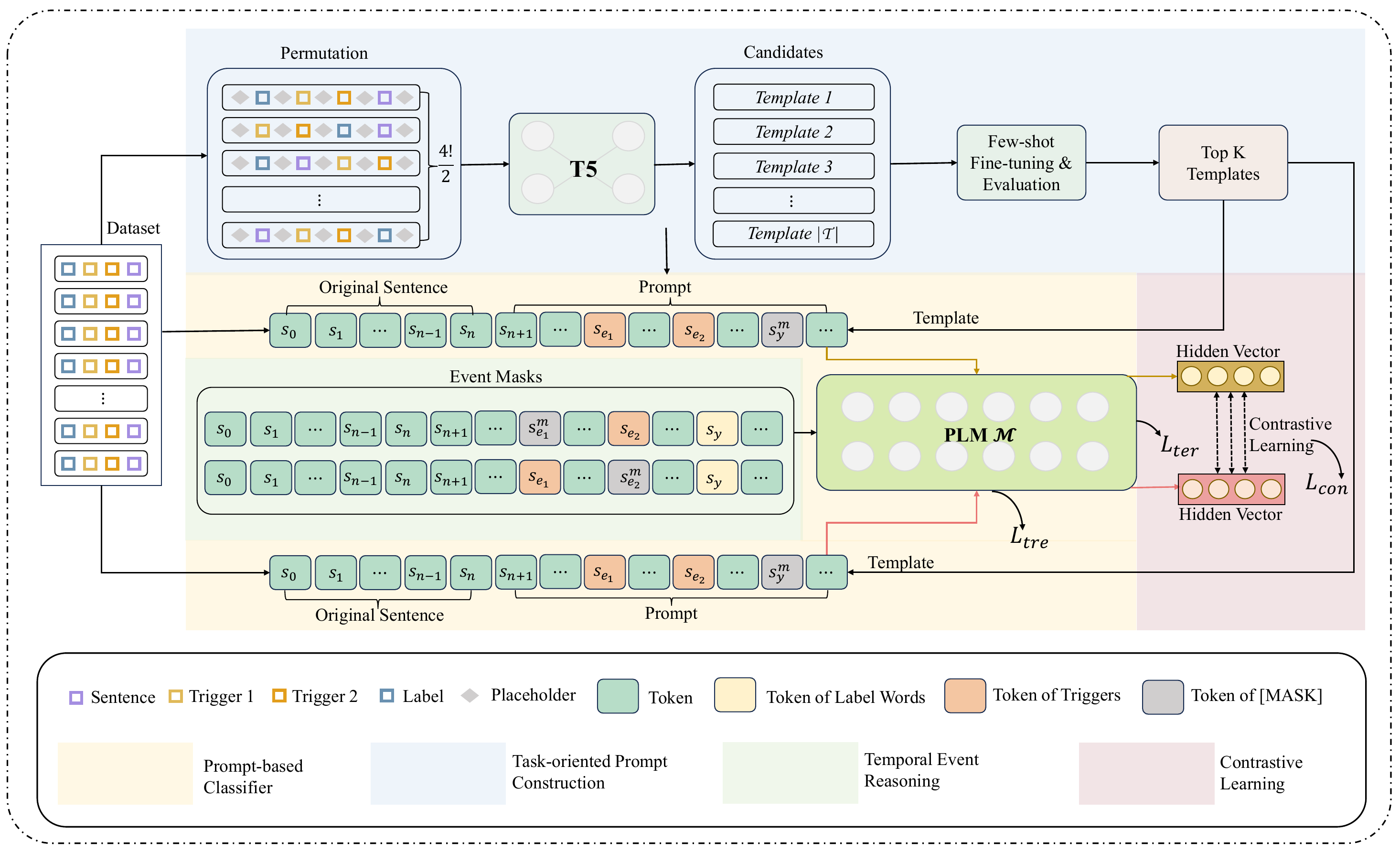}
    \caption{The overall architecture of TemPrompt, which consists of four modules: prompt-based classifier, task-oriented prompt construction, temporal event reasoning and contrastive learning.} 
    \label{fig: architecture}
\end{figure*}

\section{Problem Statement }
Given an input sentence $S_i=\{s_1, s_2,\ldots,s_n\}\in \mathcal{D}$ with two events $e_1$ and $e_2$, where $s_k$ presents the $k$-th token in $S$, $n$ is the number of tokens and $\mathcal{D}$ presents a dataset. The annotated event pairs $\langle e_1,e_2 \rangle$ correspond to the tokens $s_i$ and $s_j$ within $S$, which are specially marked as triggers $t_1$ and $t_2$. The task of TRE is to predict a relation from $\mathcal R$ between a given pair of events (e.g., $e_1$ and $e_2$ ) in a sentence $S$, where $\mathcal R=\lbrace r_i\rbrace$ is a pre-defined set of temporal relations. Note that if two annotated events originate from different sentences, those sentences are concatenated and fed as input to the model.

\section{Model Architecture}
The overall architecture of TemPrompt is illustrated in Fig. \ref{fig: architecture}, consisting of four main components: prompt-based classifier, task-oriented prompt construction, temporal event reasoning and contrastive learning. The prompt-based classifier employs prompt-based fine-tuning to stimulate the reasoning and knowledge of PLM for temporal relation classification, all without the need for additional parameters. To transform the original sentence into input for the classifier, a task-oriented prompt construction is introduced, which comprehensively considers the task-specific elements such as triggers, labels, and sentences for the automatic generation of cloze TRE prompts. Additionally, temporal event reasoning is designed as an auxiliary task to further propel the PLM's understanding of temporal clues. Contrastive learning is used to mitigate the limited amount of annotated data and its uneven distribution across relations.

\subsection{Prompt-based Classifier}
This work emphasizes the use of prompt tuning to maximize the utilization of a PLM's knowledge and capabilities while minimizing the introduction of additional parameters. To this end, we introduce a prompt-based classifier for TRE. A PLM $\mathcal{M}$ with vocabulary $Y=\lbrace{y_i}\rbrace$ is employed and a verbalizer is defined as an injective function $\mathcal{V}$: $\mathcal{R} \rightarrow Y$, which maps each label to a word from $\mathcal{M}$’s vocabulary. Each input sequence $S_i$ with triggers $t_{i1}$ and $t_{i2}$ is transformed into a masked language modeling input $S_{prompt}=\mathcal{T}(S_i,t_{i1},t_{i2},[MASK])$ through a pre-designed template $\mathcal{T}$, where $[MASK]$ refers to a token to be predicted. Subsequently, the PLM $\mathcal{M}$ receives the masked input $S_{prompt}$ and outputs the predicted probabilities $P_{\mathcal{M}}$ of relation classes $r\in \mathcal{R}$ as follows:

\begin{equation}\label{eq1}
\begin{split}
P_{\mathcal{M}}(r|S_{prompt})& =P_{\mathcal{M}}([MASK]=\mathcal{V}(r)|S_{prompt})\\
&=\frac{exp(\mathbf{w}_{\mathcal{M}(r)}\cdot \mathbf{h}_{[MASK]})}{\sum_{r^{\prime}\in \mathcal{R}}exp(\mathbf{w}_{\mathcal{M}(r^{\prime})}\cdot \mathbf{h}_{[MASK]})}
\end{split}
\end{equation}
where $\mathbf{h}_{[MASK]}$ refers to the hidden vector at the $[MASK]$ position and $\mathbf{w}_{\mathcal{M}(r^{\prime})}$ denotes the pre-softmax vector corresponding to $y\in Y$. 

To achieve more accurate predictions, the supervised samples $\lbrace(S_i,r_i)\rbrace$ are constructed and utilized to fine-tune the PLM $\mathcal{M}$ by minimizing the cross-entropy loss $\mathcal{L}_{tre}$:

\begin{equation}\label{eq1}
\mathcal{L}_{tre}=-\sum_{i=0}^{\lvert \mathcal{R} \rvert-1}P^{gt}(r_i|S_{prompt})log P_{\mathcal{M}}(r_i|S_{prompt})
\end{equation}
where $P^{gt}$ is a ground-truth probability distribution of temporal relations.

\subsection{Task-oriented Prompt Construction}
We introduce a TRE task-oriented prompt construction approach for automatically creating a variety of cloze prompts, which takes sentences, labels and pairs of triggers into consideration. Specifically, given an input quadruple $x_i=(S_i,t_{i1}, t_{i2}, y_i)\in \mathcal{D}_{train}$, we permute its elements according to the permutation function $f_p$ as follows: 
\begin{equation}\label{eq1}
f_p(S_i,t_{i1}, t_{i2}, y_i)=
\begin{Bmatrix}
(S_i, y_i, t_{i1}, t_{i2})\\
(y_i, S_i, t_{i1}, t_{i2})\\
(S_i, t_{i1}, y_i, t_{i2})\\
(y_i, t_{i1}, S_i, t_{i2})\\
(t_{i1}, S_i, y_i, t_{i2})\\
(t_{i1}, y_i, S_i, t_{i2})\\
(t_{i1}, t_{i2}, y_i, S_i)\\
(t_{i1}, t_{i2}, S_i,y_i)\\
(t_{i1}, S_i, t_{i2}, y_i)\\
(t_{i1}, y_i, t_{i2}, S_i)\\
(S_i, t_{i1}, t_{i2}, y_i) \\ 
(y_i, t_{i1}, t_{i2}, S_i)
\end{Bmatrix}
\end{equation}
where $4!/2=\frac{4*3*2*1}{2}=12$ permutations should be considered as two triggers $t_{i1}$ and $t_{i2}$ may not be adjacent, yet their order should remain constant for accurate label matching. 


Therefore, templates $\mathcal{T}(S_i,t_{i1}, t_{i2}, y_i)$ are constructed following the format:  \begin{equation}\label{eq1}
\mathcal{T}(S_i,t_{i1}, t_{i2}, y_i)=
\begin{Bmatrix}
(\langle X \rangle S_i \langle Y \rangle y_i \langle Z \rangle t_{i1} \langle M \rangle t_{i2} \langle N \rangle)\\
(\langle X \rangle y_i \langle Y \rangle S_i \langle Z \rangle t_{i1} \langle M \rangle t_{i2} \langle N \rangle)\\
(\langle X \rangle S_i \langle Y \rangle t_{i1} \langle Z \rangle  y_i \langle M \rangle t_{i2} \langle N \rangle)\\
(\langle X \rangle y_i \langle Y \rangle t_{i1} \langle Z \rangle S_i \langle M \rangle t_{i2} \langle N \rangle)\\
(\langle X \rangle t_{i1} \langle Y \rangle S_i \langle Z \rangle y_i \langle M \rangle t_{i2} \langle N \rangle)\\
(\langle X \rangle t_{i1} \langle Y \rangle y_i \langle Z \rangle S_i \langle M \rangle t_{i2} \langle N \rangle)\\
(\langle X \rangle t_{i1} \langle Y \rangle t_{i2} \langle Z \rangle y_i \langle M \rangle S_i \langle N \rangle)\\
(\langle X \rangle t_{i1} \langle Y \rangle t_{i2} \langle Z \rangle S_i \langle M \rangle y_i \langle N \rangle)\\
(\langle X \rangle t_{i1} \langle Y \rangle S_i \langle Z \rangle t_{i2} \langle M \rangle y_i \langle N \rangle)\\
(\langle X \rangle t_{i1} \langle Y \rangle y_i \langle Z \rangle t_{i2} \langle M \rangle S_i \langle N \rangle)\\
(\langle X \rangle S_i \langle Y \rangle t_{i1} \langle Z \rangle t_{i2}\langle M \rangle y_i \langle N \rangle) \\
(\langle X \rangle y_i \langle Y \rangle t_{i1} \langle Z \rangle t_{i2}\langle M \rangle S_i \langle N \rangle)\\

\end{Bmatrix}
\end{equation}
where $\langle X \rangle$, $\langle Y \rangle$, $\langle Z \rangle$, $\langle M \rangle$ and $\langle N \rangle$ are placeholders to be filled in with concrete words. The original input elements serve as predefined constraints that shape the generation of prompts. In this paper, we select and employ T5 \cite{raffel2020exploring}, a pre-trained text-to-text transformer, to supplement missing texts within placeholders, as it is trained to predict masked spans.  

We leverage beam search with a wide beam width (e.g., 200) to decode a diverse pool of candidate templates. Our decoding goal is to maximize the joint probability $P_{joint}$ of template tokens $(s_1,s_2,\ldots,s_{|\mathcal{T}|})$, which can facilitate the creation of a template that generalizes well to all training examples in $\mathcal{D}_{train}$. 

\begin{equation}\label{eq1}
P_{joint}=\sum_{j=1}^{\lvert \mathcal{T} \vert}\sum_{x_i \in \mathcal{D}_{train}} logP_{T5}(s_j|s_1,s_2,\ldots, s_{j-1}, \mathcal{T}(x_i))
\end{equation}
where $P_{T5}$ presents the output probability distribution predicted by T5 and $|\mathcal{T}|$ is the length of a template $\mathcal{T}$. 

After generating a diverse set of candidate templates, we fine-tune a PLM $\mathcal{M}$ for TRE based on them on $\mathcal{D}_{train}$ and evaluate their performance on $\mathcal{D}_{dev}$ using the F1 score. The top K templates $TopK$ with the highest F1 scores are selected and then combined with elements $S_i$, $t_{i1}$ and $t_{i2}$ to construct prompts as follows:

\begin{equation}
\begin{split}
TopK= & argmax_{{\lbrace \mathcal{T}_k \rbrace}_K}\lbrace F1( P_{\mathcal{M}}([MASK]=\mathcal{V}(r_i)| \\
& \mathcal{T}_k(S_i,t_{i1},t_{i2},[MASK])), x_i \in \mathcal{D}_{dev})\rbrace
\end{split}
\end{equation}
where $P_{\mathcal{M}}$ is the output probability distribution of $\mathcal{M}$. $\mathcal{D}_{train}$ and $\mathcal{D}_{dev}$ are small-scale datasets sampled from the original training set, thereby expediting the fine-tuning process. 
\subsection{Temporal Event Reasoning}

TRE models place a strong emphasis on extracting relevant information from event mentions while contexts serve as the primary source for supporting accurate predictions. However, because vanilla PLM gives equal treatment to all tokens, it often fails to adequately capture the nuances of temporal event knowledge embedded in the context. Considering that both the input and output spaces are comprised of tokens, we can explore an alternative perspective: whether a PLM can identify another event when provided with one event and a temporal relation. Therefore, to achieve this, we design a temporal event reasoning task as an auxidiary task to fine-tune PLMs for understanding an event via considering its temporal relations with other events in the text. 


\begin{figure}[!h]
    \centering  
    \includegraphics[width=1\linewidth]{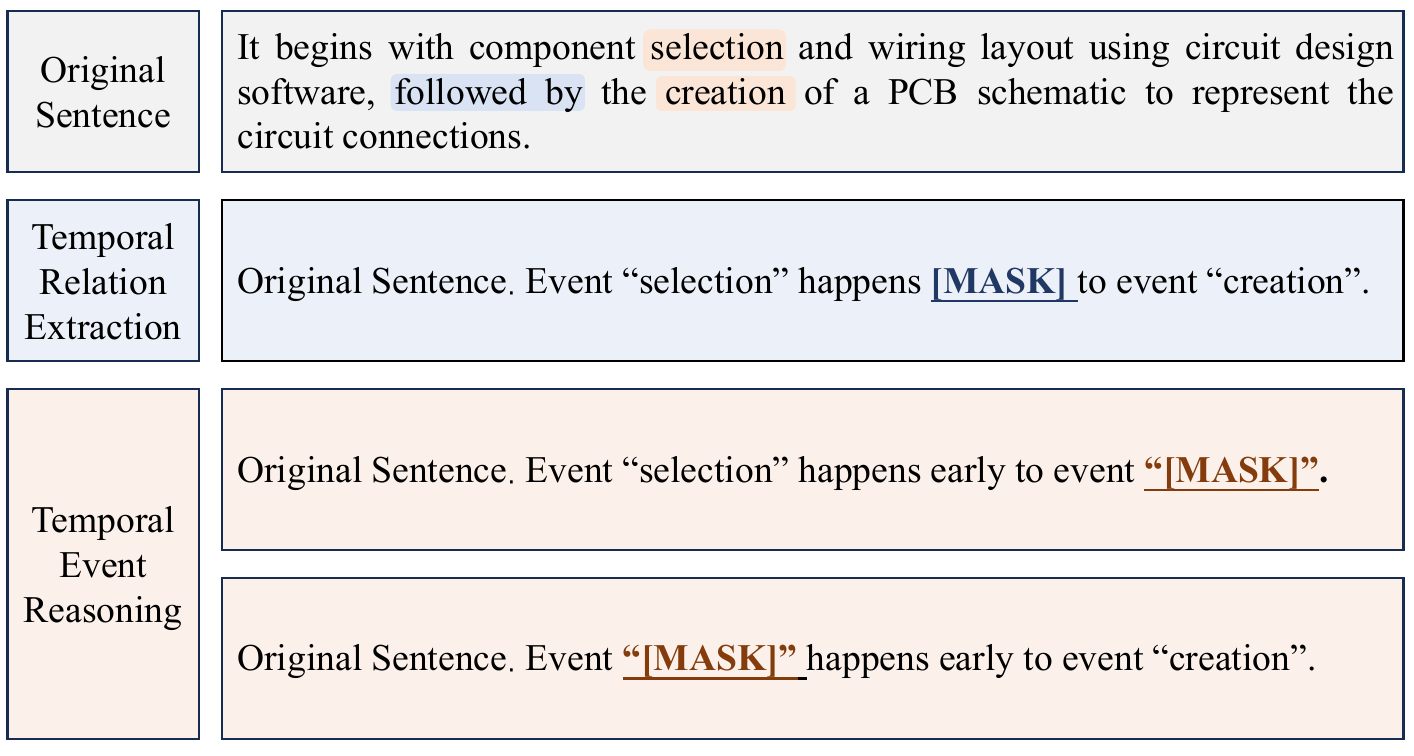}
    \caption{Masking instances of auxiliary tasks.} 
    \label{fig: auxiliary}
\end{figure}

Specifically, as shown in Fig. \ref{fig: auxiliary}, we substitute event triggers (e.g., selection and creation) with $[MASK]$ tokens, that is, $S_{prompt1}=\mathcal{T}(S_i,[MASK],t_{i2}, y_i)$ and $S_{prompt2}=\mathcal{T}(S_i,t_{i1},[MASK], y_i)$. For example, we create “\textit{Original Sentence. Event “$\mathit{[MASK]}$” happens early to event “creation}” or “\textit{Original Sentence. Event “selection” happens early to event “$\mathit{[MASK]}$}” as the input of PLMs’s auxiliary task. Obviously, these two prompts hold the promise of enhancing PLMs' attention to events and facilitating a better understanding of interactions between events and their temporal relations. Given an input sequence $S=\lbrace {s_1,s_2, \ldots,s_n}\rbrace$, it is masked as $S_{prompt1}$ or $S_{prompt2} = S^m=\lbrace {s_1, \ldots, s_j^m,\ldots, s_n}\rbrace$, where $s_j^m$ denotes the masked tokens of an event ($t_{i1}$ or $t_{i2}$) trigger. The objective of auxiliary tasks is to predict the identity $s_j$ of $s_j^m$ by minimizing a cross-entropy loss $\mathcal{L}_{ter}$:



\begin{equation}
\begin{split}
\mathcal{L}_{ter}= -\frac{1}{2} &\lbrace \sum_{s_j^m \in S}  \mathcal{I}[s_j^m=s_j]logP_{\mathcal{M}}(s_j^m|S_{prompt1})\\
& +\sum_{s_j^m \in S} \mathcal{I}[s_j^m=s_j]logP_{\mathcal{M}}(s_j^m|S_{prompt2})\rbrace
\end{split}
\end{equation}
where $\mathcal{I}[\cdot]$ is an indicator function that returns 1 if the condition is true, and 0 otherwise. 

\subsection{Contrastive Learning}
To alleviate the uneven relation distribution within the dataset, we employ contrastive learning to pull representations of similar samples closer together and push those of dissimilar samples farther apart. Specifically, we divide the training set into several batches $\mathcal{D}_b=\lbrace{S_i}\rbrace, i=0,1,2,\ldots, m-1$, where $m$ refers to batch size. For a given input sample $S_i$, we can identify similar samples of the same relation category $N(S_i)=\lbrace{S_j|r_j=r_i}\rbrace$. Then, we further optimize the PLM $\mathcal{M}$ by minimizing the contrastive loss $\mathcal{L}_{con}$: 

\begin{equation}
\begin{split}
& \mathcal{L}_{con}=\\ & \sum_{S_i \in \mathcal{D}_b}\frac{-1}{\lvert N(S_i) \rvert}\sum_{S_j \in N(S_i)} log(\frac{exp((\mathbf{h}_{S_i}\cdot\mathbf{h}_{S_j})/\tau)}{\sum_{S_k\in \mathcal{D}_b \backslash \lbrace S_i \rbrace}exp((\mathbf{h}_{S_k}\cdot\mathbf{h}_{S_i})/\tau)})
\end{split}
\end{equation}
where $\mathbf{h}_{S_i}$ refers to the hidden vector of $[MASK]$ within the $S_i$-related prompt, ${\lvert N(S_i) \rvert}$ denotes the number of similar samples to $S_i$, $\mathcal{D}_b \backslash \lbrace S_i \rbrace$ represents the removal of the input $S_i$ from the set $\mathcal{D}_b$ and $\tau$ is a temperature coefficient used to adjust the value magnitude. 

\subsection{The Total Loss}
In summary, the total loss is the summation of classification loss of temporal relations, language modeling loss of auxiliary tasks and contrastive loss, as follows:
\begin{equation}
\mathcal{L}_{total}=\mathcal{L}_{tre}+\alpha*\mathcal{L}_{ter}+\beta*\mathcal{L}_{con}
\end{equation}
where $\alpha$ and $\beta$ are trade-off hyper-parameters among three types of losses.

\section{Experiments}
\subsection{Datasets}
We conduct a series of experiments on two public benchmark datasets for TRE: TB-Dense \cite{cassidy2014annotation} and MATRES\cite{ning2018multi}. TB-Dense is based on TimeBank Corpus \cite{pustejovsky2003timebank} and annotated densely with six types of relations: \textit{BEFORE}, \textit{AFTER}, \textit{SIMULTANEOUS}, \textit{INCLUDES}, \textit{INCLUDED IN} and \textit{VAGUE}.  It requires annotating all event/time pairs within a specified window and introduces a \textit{VAGUE} label to capture difficult-to-distinguish relations even by humans.  MATRES further simplifies temporal relations into four distinct categories: \textit{BEFORE}, \textit{AFTER}, \textit{EQUAL} and \textit{VAGUE}. It offers refined TRE annotations on TimeBank \cite{pustejovsky2003timebank} and TempEval \cite{uzzaman2013semeval} (including AQUAINT and Platinum subsets) documents. Our experimental setup for both benchmark datasets involves utilizing the same training/validation/testing splits as those used in previous studies \cite{wen2021utilizing,han2019deep,zhuang2023knowledge}. Their detailed statistics and data split can be found in Table \ref{dataset}. 

\begin{table}[t]
  \centering
  \caption{Statistics of TB-Dense and MATRES datasets.}
  \label{dataset}
  \setlength{\tabcolsep}{2mm}{
  \resizebox{1\linewidth}{!}{
  \begin{tabular}{llccc}
    \toprule
    Dataset& & Training & Validation & Test\\
    \midrule
    \multirow{2}{*}{MATRES} & Documents & 260 & 21& 20\\
    &Event Pairs & 10,888 & 1,852 & 840\\
    \midrule
    \multirow{2}{*}{TB-Dense} & Document & 22 & 5 & 9\\
    & Event Pairs & 4,032 & 629 & 1,427\\
    \bottomrule
  \end{tabular}
}}
\end{table}

\subsection{Baselines}
We compare our model with a series of different representative baseline models on different datasets. We utilize the reported results in their original paper and populate the table with a “\text{-}” symbol for Precision and Recall values that are not provided.

For the MATRES dataset, the following baselines are chosen for comparison. \textbf{(1) TEMPROB+ILP} \cite{ning2019improved}: A method that injects knowledge from TEMPROB, a temporal common sense knowledge base, and achieves global inference via integer linear programming. \textbf{(2) Joint Constrain} \cite{wang2020joint}: A method that incorporates the contextualized representations along with statistical common-sense knowledge from several knowledge bases and defines declarative logic rules for joint constrained learning. \textbf{(3) Vanilla Classifier} \cite{wen2021utilizing}: A method that inputs the concatenated  pre-trained representations of given two event mentions into a feed-forward neural network for classification. \textbf{(4) Self-training} \cite{ballesteros2020severing}: A method that leverages multi-task learning and self-training techniques based on multiple complementary datasets. \textbf{(5) Relative Time} \cite{wen2021utilizing}: A method adopts a stack-propagation framework to incorporate relative event time prediction as an auxiliary task. \textbf{(6) Probabilistic Box} \cite{hwang2022event}: A method that models asymmetric relationships between entities by projecting each event to a box representation. \textbf{(7) Syntax Transformer} \cite{zhang2021extracting}: A method that explicitly represents the connections between two events by building a dependency graph and leverages a novel syntax-guided attention mechanism to explore local temporal cues. \textbf{(8) OntoEnhance} \cite{zhuang2023knowledge}: A method that injects ontology knowledge into text prompts via event type prediction and introduces a dual-stack attention fusion mechanism to alleviate knowledge noise. Additionally, we compare using random masking instead of temporal event reasoning as \textbf{TemPrompt (Rand)} with our original model \textbf{TemPrompt (TRE)}.

We assess our model's performance relative to the chosen baselines on TB-Dense dataset, where \textbf{(4) Vanilla Classifier} \cite{wen2021utilizing}, \textbf{(6) Syntax Transformer} \cite{zhang2021extracting} and \textbf{(8) OntoEnhance} \cite{zhuang2023knowledge} are shared with MATRES, and others are as follows. \textbf{(1) Deep Structured} \cite{han2019deep}: A method that incorporates domain knowledge via a structured support vector machine and learns scoring functions for pair-wise relations via a RNN. \textbf{(2) SEC}: A method that builds an event centric model across three TLINK categories (E2E, E2T, and E2D) with multi-task learning. \textbf{(3) Uncertainty-training} \cite{cao2021uncertainty}: A method that utilizes uncertainty-aware self-training to cope with pseudo-labeling errors in TRE. \textbf{(5) PSL} \cite{zhou2021clinical}: A method that formulates the probabilistic soft logic rules of temporal dependencies as a regularization term and globally infers the temporal relations with the time graphs. \textbf{(7) ECONET} \cite{han2020econet}: A method that pre-trains a PLM with the objective of recovering masked-out event and temporal indicators and discriminating sentences from their corrupted counterparts. Similar to MATRES, we also test \textbf{TemPrompt (Rand)} and \textbf{TemPrompt (TRE)} models. 

\subsection{Training Settings and Evaluation Metrics}
To be consistent with the previous work\cite{han2020econet, wen2021utilizing,zhang2021extracting,zhuang2023knowledge}, our TemPrompt model is implemented based on the RoBERTa-Large as the PLM $\mathcal M$. We fine-tune it with a batch size of 16 for 10 epochs. We use the AdamW optimizer with a learning rate of $5e-5$ and a linear scheduler for the smooth optimization of the model's parameters. In terms of task-oriented prompt construction, we configure the beam width to 200 and finally select the top $K=5$ templates for evaluation.
For contrastive learning, the temperature coefficient $\tau$ is set to 0.2. The trade-off hyperparameters $\alpha$ and $\beta$ for the total loss are configured as 1 and 0.5, respectively. The standard precision (P), Recall (R) and micro-average F1 (F1) metrics are utilized as the evaluation criteria of models' performance. All experiments are performed on a Linux server with eight Tesla V100 GPUs.

\begin{table}[t]
  \centering
  \caption{Results of various TRE methods on MATRES.}
  \label{Res_MATRES}
  \setlength{\tabcolsep}{3.5mm}{
  \resizebox{1\linewidth}{!}{
  \begin{tabular}{l|ccc}
    \toprule
    Model & P & R & F1\\
    \midrule
    TEMPROB+ILP & 71.3 & 82.1 & 76.3\\
    Joint Constrain & 73.4 & 85.0 & 78.3\\
    Vanilla Classifier & 78.1 & 82.5 & 80.2\\
    Self-Training & \text{-} & \text{-} & 81.6\\ 
    Relative Time & 78.4 & \underline{85.2} & 81.7\\
    Probabilistic Box & \text{-} & \text{-} & 77.1\\
    Syntax Transformer & \text{-} & \text{-} & 80.3 \\
    OntoEnhance & \underline{79.0} & \textbf{86.5} & \underline{82.6}\\
    \midrule
    TemPrompt (Rand) & 78.5 & 84.3 & 81.3 \\
    TemPrompt (TRE) & \textbf{82.0} & 83.9 & \textbf{82.9}\\
    \bottomrule
  \end{tabular}
}}
\end{table}

\begin{table}[t]
  \centering
  \caption{Results of various TRE methods on TB-Dense.}
  \label{Res_TB}
  \setlength{\tabcolsep}{3.5mm}{
  \resizebox{1\linewidth}{!}{
  \begin{tabular}{l|ccc}
    \toprule
    Model & P & R & F1\\
    \midrule
    Deep Structured & \text{-} & \text{-} & 63.2\\
    SEC & \text{-} & \text{-} & 65.0\\
    Uncertainty-training & 60.8 & 60.8 & 60.8 \\
    Vanilla Classifier &  62.1 &  64.0 & 63.0 \\
    PSL & \text{-} & \text{-} &  65.2\\  
    Syntax Transformer & \text{-} & \text{-} & 67.1 \\
    ECONET & \text{-} & \text{-} &  66.8\\
    OntoEnhance &  67.5  & 68.6 & 68.0 \\
    \midrule
    TemPrompt (Rand) & \underline{69.3} & \underline{70.2} & \underline{69.7} \\
    TemPrompt (TRE) & \textbf{70.7} & \textbf{71.3} & \textbf{71.0}\\
    \bottomrule
  \end{tabular}
}}
\end{table}

\subsection{Results and analysis}
The results of our models and various baselines on the MATRES and TB-Dense datasets are shown in Table \ref{Res_MATRES} and \ref{Res_TB}, with the best results highlighted in bold and the second-best results underlined. Except for the R score on MATRES, our approach outperforms all baseline models on the two datasets. Specifically, our TemPrompt (TRE) model achieves improvements of 0.3\% and 3\% in F1 scores on MATRES and TB-Dense, respectively, compared to state-of-the-art models. This demonstrates the effectiveness and superiority of our method in extracting temporal relations.



Our TemPrompt (TRE) model exhibits better performance than a series of external knowledge-based methods (i.e., TEMPROB+ILP, Joint Constrain, Probabilistic Box, Syntax Transformer, Deep Structured, PSL and OntoEnhance), involving common-sense knowledge base, dependency graphs and declarative logic rules. This suggests that the introduction of external knowledge is both cumbersome and of limited effectiveness, as comparable performance can be achieved solely through the exploration of the internal knowledge stored within the PLMs. This could be attributed to the formidable challenges that this method encounters, such as knowledge missing, knowledge noise and suboptimal knowledge injection. It also outperforms all the data augmentation baselines (i.e., self-training, SEC and Uncertainty-training), indicating that it can achieve exceptional TRE results without the necessity of additional datasets. Especially, its superior performance over Relative Time validates the effectiveness of our auxiliary task design.


TemPrompt (Rand) improves the F1 scores by 1.1\% and 8\% compared to Vanilla Classifier on MATRES and TB-Dense. Therefore, its superior performance across all metrics in contrast to the Vanilla Classifier highlights the enhanced utilization of PLM’s reasoning and generation capabilities through fine-tuning. Furthermore, its performance surpassing ECONET on TB-Dense indicates that prompts play a crucial role in better activating the PLM's knowledge. However, it demonstrates worse performance than TemPrompt (TRE). I believe this is attributed to the capability of temporal event reasoning, as an auxiliary task, to more effectively mine temporal clues.

\begin{table}[t]
  \centering
  \caption{Ablation study on MATRES and TB-Dense.}
  \label{ablation}
  \setlength{\tabcolsep}{3.5mm}{
  \resizebox{1\linewidth}{!}{
  \begin{tabular}{c|l|ccc}
    \toprule
    Dataset & Model & P & R & F1\\
    \midrule
    \multirow{3}{*}{MATRES} & TemPrompt & \textbf{82.01} & \textbf{83.93} & \textbf{82.96}\\
    & -wo TER & 81.83 & 82.28 & 82.05\\
    & -wo CL & 80.24 & 83.10 & 81.65\\
    \midrule
     \multirow{3}{*}{TB-Dense}& TemPrompt & \textbf{70.71} & \textbf{71.32} & \textbf{71.01}\\
      & -wo TER &  69.96 & 70.67 & 70.31\\
    & -wo CL &  70.45 & 70.96 & 70.70\\
    \bottomrule
  \end{tabular}
}}
\end{table}

\subsection{Ablation Study}
To evaluate the impact of each component on the overall performance of the TemPrompt model, we conduct ablation experiments separately on two datasets. We remove temporal event reasoning and contrastive learning individually, denoted as \textbf{-wo TER} and \textbf{-wo CL}, for testing purposes. The results of ablation studies are shown in Table \ref{ablation}. After removing both, we observe a decline in metrics, highlighting the significant positive influence of both on the performance of TRE. However, the extent of performance decline varies across different datasets. The removal of temporal event reasoning leads to a decrease of 0.91\% in F1 score on MATRES and a decrease of 0.7\% on TB-Dense, while the absence of contrastive learning resulted in a 1.31 \% decrease in F1 score on MATRES and a 0.31\% decrease on TB-Dense. This might be attributed to the fact that in TB-Dense, two events unfold within continuous lengthy texts, yet we only merge sentences containing these two events. This disruption of contextual semantic continuity may impede overall contextual comprehension and consequently weaken the promotional effect of temporal event reasoning.

\begin{table*}[t]
  \centering
  \caption{Results of different templates on MATRES.}
  \label{prompt}
  \setlength{\tabcolsep}{3.5mm}{
  \resizebox{1\linewidth}{!}{
  \begin{tabular}{c|c|ccc}
    \toprule
    Name & Template & P & R & F1\\
    \midrule
    \multicolumn{5}{c}{Mannully}\\
    \midrule
    Prompt0 &  ‘sentence’ The temporal relation between ‘event1’ and ‘event2’ is $\langle$mask$\rangle$. & 80.17 & 78.85 & 79.50 \\
    \midrule
    \multicolumn{5}{c}{Automatically}\\
    \midrule
    Prompt1 &  ‘sentence’ The word ‘event1’ and the word ‘event2’ are used $\langle$mask$\rangle$ in the sentence. & 77.19 & 82.28 & 79.65 \\
    Prompt2 & ‘sentence’ Note that the words ‘event1’ and ‘event2’ are used $\langle$mask$\rangle$ in the sentence. & 78.93 & 83.38 & 81.10\\
    Prompt3 &  ‘sentence’ The words ‘event1’ and ‘event2’ were $\langle$mask$\rangle$ used in the same sentence. & 79.52 & 82.69 & 81.08 \\
    Prompt4 &  ‘sentence’ The word ‘event1’ is used $\langle$mask$\rangle$ in the article. It means ‘event2’. & 80.32 & 81.87 & 81.09\\
    Prompt5 &  ‘sentence’ Event ‘event1’ happened $\langle$mask$\rangle$ to ‘event2’. & \textbf{82.01} & \textbf{83.93} & \textbf{82.96} \\
    \bottomrule
  \end{tabular}
}}
\end{table*}

\subsection{Few-shot Event Study}

\begin{figure}[htbp]
\centering 
\subfigure[MATRES]{
\begin{minipage}[t]{0.48\linewidth}
\centering
\includegraphics[width=1.8in]{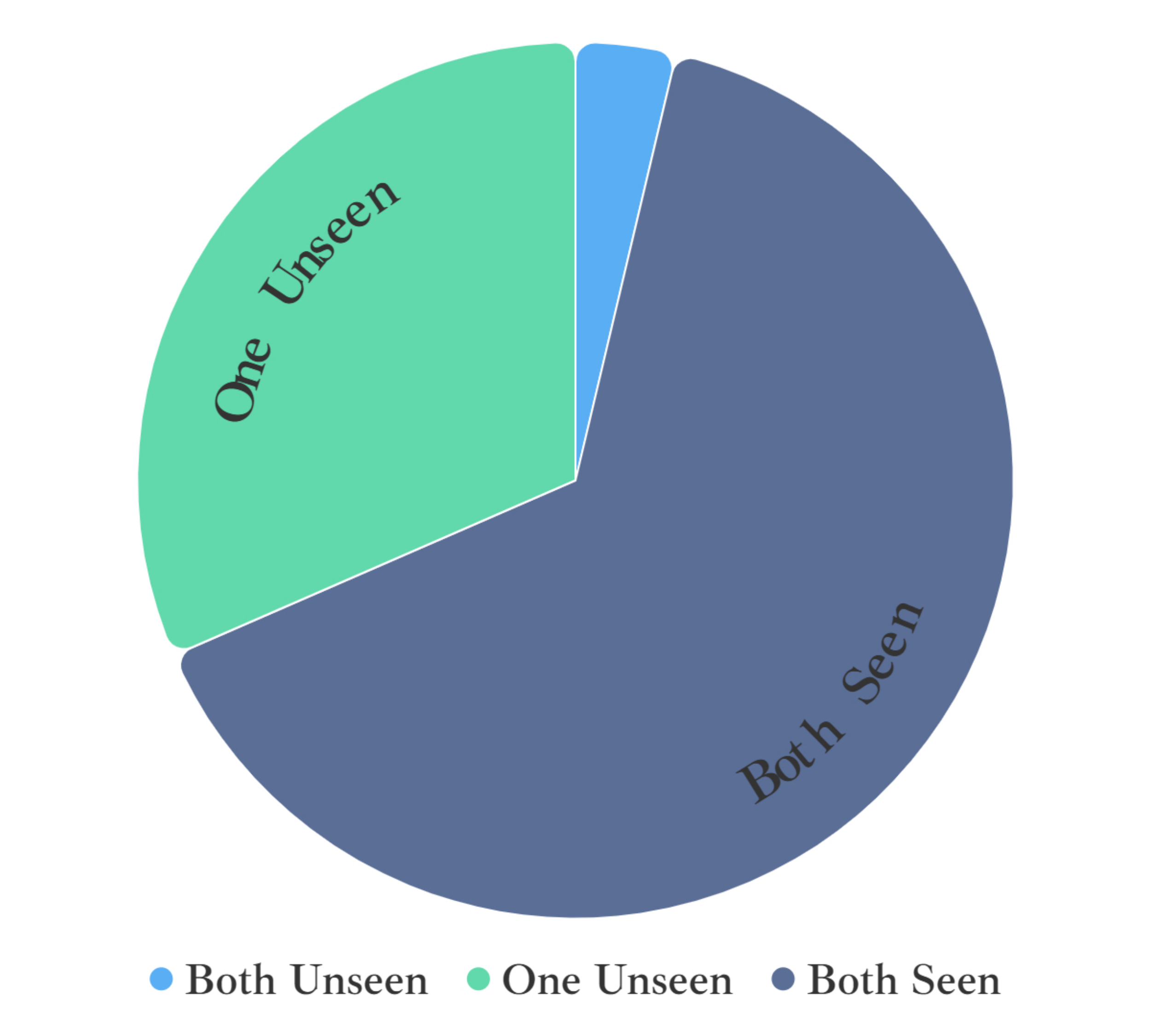}
\end{minipage}%
}%
\subfigure[TB-Dense]{
\begin{minipage}[t]{0.48\linewidth}
\centering
\includegraphics[width=1.8in]{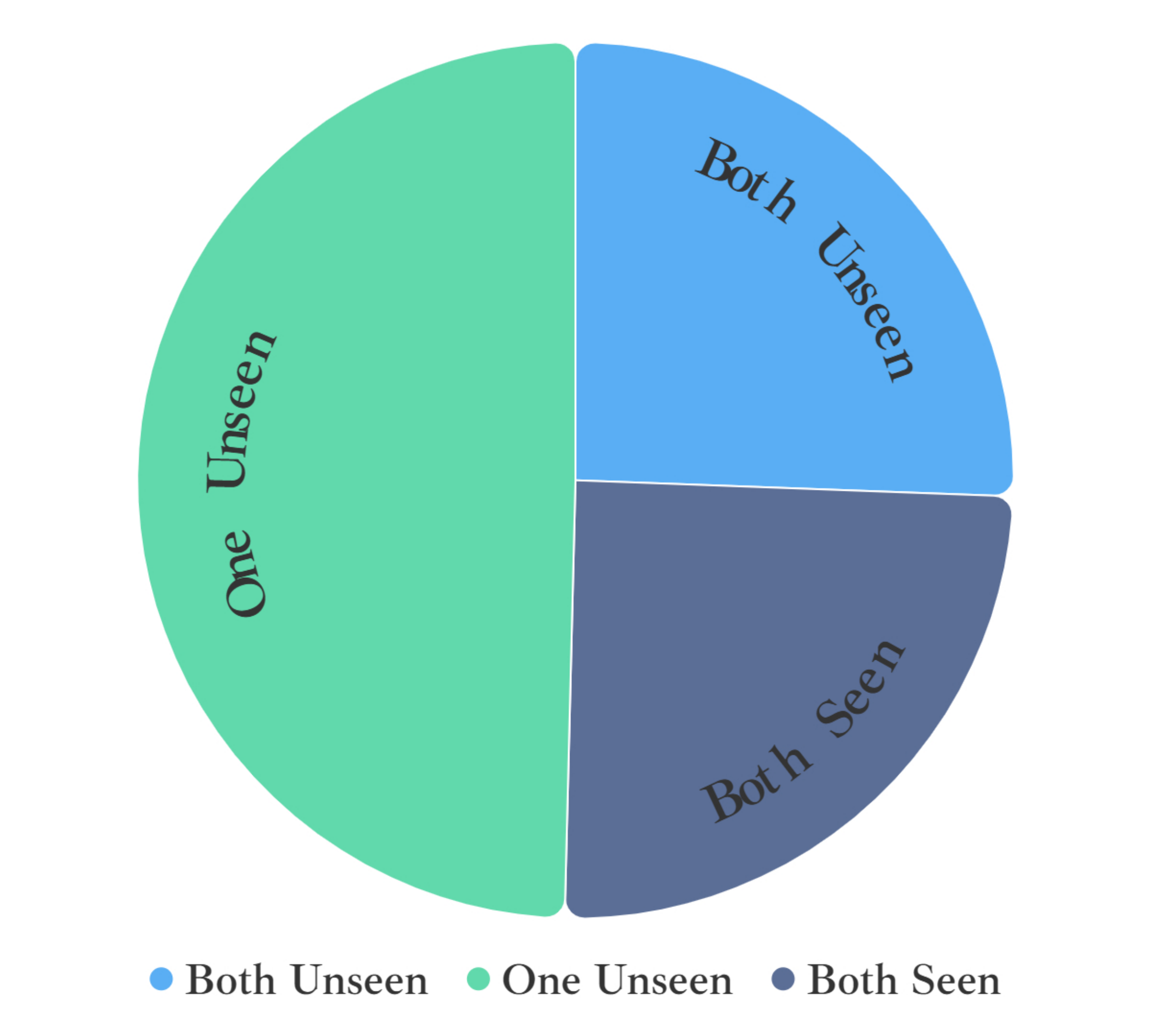}
\end{minipage}%
}%
\centering
\caption{The proportion of samples across three event categories in the testing sets.}
\label{Few-shot}
\end{figure}

\begin{figure}[htbp]
\centering 
\subfigure[MATRES]{
\begin{minipage}[t]{0.48\linewidth}
\centering
\includegraphics[width=1.8in]{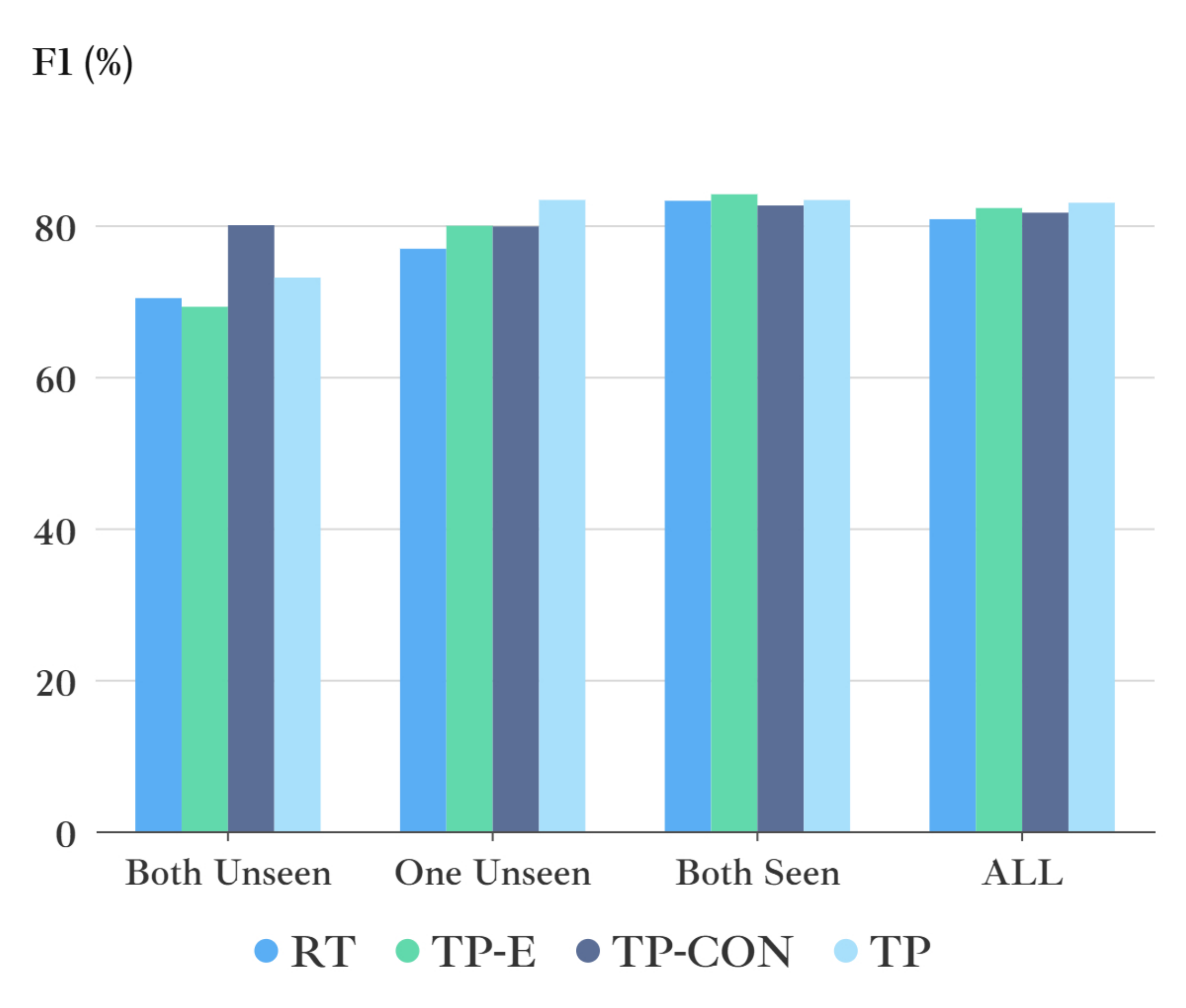}
\end{minipage}%
}%
\subfigure[TB-Dense]{
\begin{minipage}[t]{0.48\linewidth}
\centering
\includegraphics[width=1.8in]{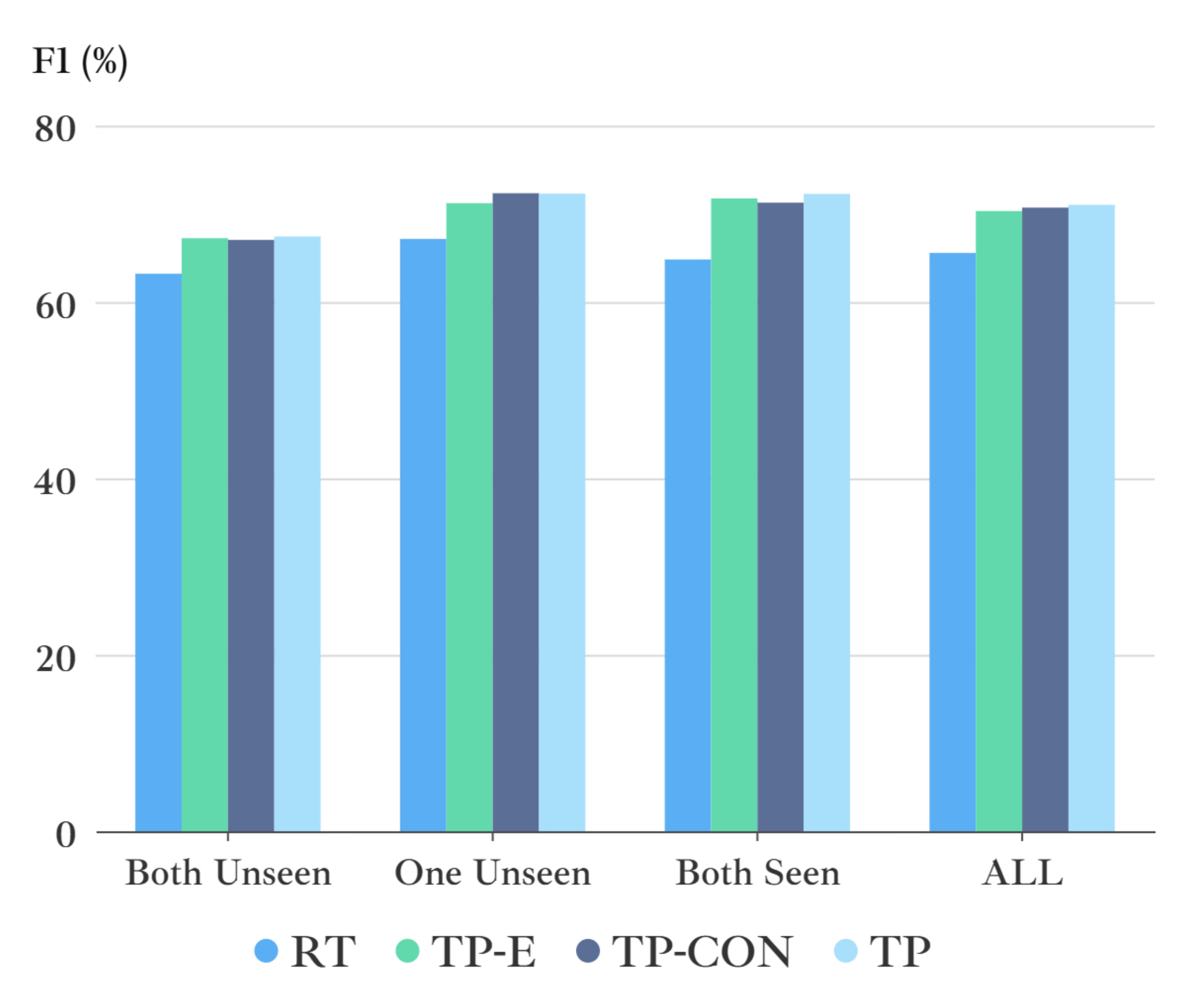}
\end{minipage}%
}%
\centering
\caption{Results of “Both Unseen”, “One Unseen” and “Both Seen” events on MATRES and TB-Dense.}
\label{fewResults}
\end{figure}

To further evaluate OntoEnhance's ability to generalize for few-shot events, we categorized the testing sets of MATRES and TB-Dense into the following three groups, as shown in Fig. \ref{Few-shot}: \textbf{(1) Both Unseen:} Neither of the events in the event pair are present in the training sets, with respective proportions of approximately 4\% on MATRES and 26\% on TB-Dense. \textbf{(2) One Unseen:} In the event pair, only one event can be found in the training sets, accounting for approximately 32\% on MATRES and 50\% on TB-Dense. \textbf{(3) Both Seen:} Both events in the event pair are included in the training sets and their occurrence is approximately 5\% on MATRES and 8\% on TB-Dense. Four distinct models are chosen to compare and analyze data across three categories, as demonstrated in Fig. \ref{fewResults}. \textbf{TP} stands for the complete TemPrompt model that we propose. \textbf{TP-CON} represents the TemPrompt model lacking contrastive learning, whereas \textbf{TP-E} indicates the TemPrompt model without temporal event reasoning. \textbf{RT} refers to the Relative Time model with open-source codes, which regards relative time prediction as an auxiliary task. 

Obviously, compared to RT, TP and its variants demonstrate superior performance in few-shot event scenarios, which provides further evidence of the effectiveness of prompt-based fine-tuning in facilitating PLMs' comprehension of the temporal evolution implied within contexts. Among them, in terms of both unseen events, TP-CON achieves a 9.63\% improvement in F1 score on MATRES and a 3.83\% improvement on TB-Dense, highlighting the substantial performance enhancement of TRE through the auxiliary task of temporal event reasoning in mining the temporal cues of few-shot and even zero-shot events. However, TP results in a 6.92\% decrease in F1 compared to TP-CON on the MATRES dataset, indicating that contrastive learning may lead to overfitting and consequently diminish the model's generalization. Overall, our TemPrompt model achieves the best performance across all the data, striking a balance between few-shot and standard settings.

\subsection{Analysis of Different Templates}
To assess the interpretability and effectiveness of various prompts, we examine the top five generated templates (i.e., Prompt1, Prompt2, Prompt3, Prompt4 and Prompt5) that demonstrate the best performance by TemPrompt on the small-scale dataset, as well as manually designed templates (i.e., Prompt0). Based on these templates, we evaluate and compare their results on the overall test set of MATRES, as shown in Table \ref{prompt}. It is observed that the automatically generated templates exhibit grammatical accuracy, and coherent phrasing, and are easily interpretable. Additionally, they achieve higher F1 scores than manual templates. This can be attributed to the tendency for manually created templates to be suboptimal when attempting to elicit the correct response from the language model in tasks involving multiple categories. The model's performance is sensitive to template selection, as illustrated by the 3.31\% higher F1 score achieved by Prompt5 compared to Prompt1, despite both being automated templates. Therefore, our few-shot evaluation strategy is important and indispensable. This further exemplifies the superiority of our task-oriented prompt generation. Despite their apparent interpretability to humans, manually crafted templates often fail to align with the language model's interpretation of inputs, thereby compromising model performance.

\begin{figure}[htbp]
\centering 
\subfigure[Before]{
\begin{minipage}[t]{0.48\linewidth}
\centering
\includegraphics[width=1.7in]{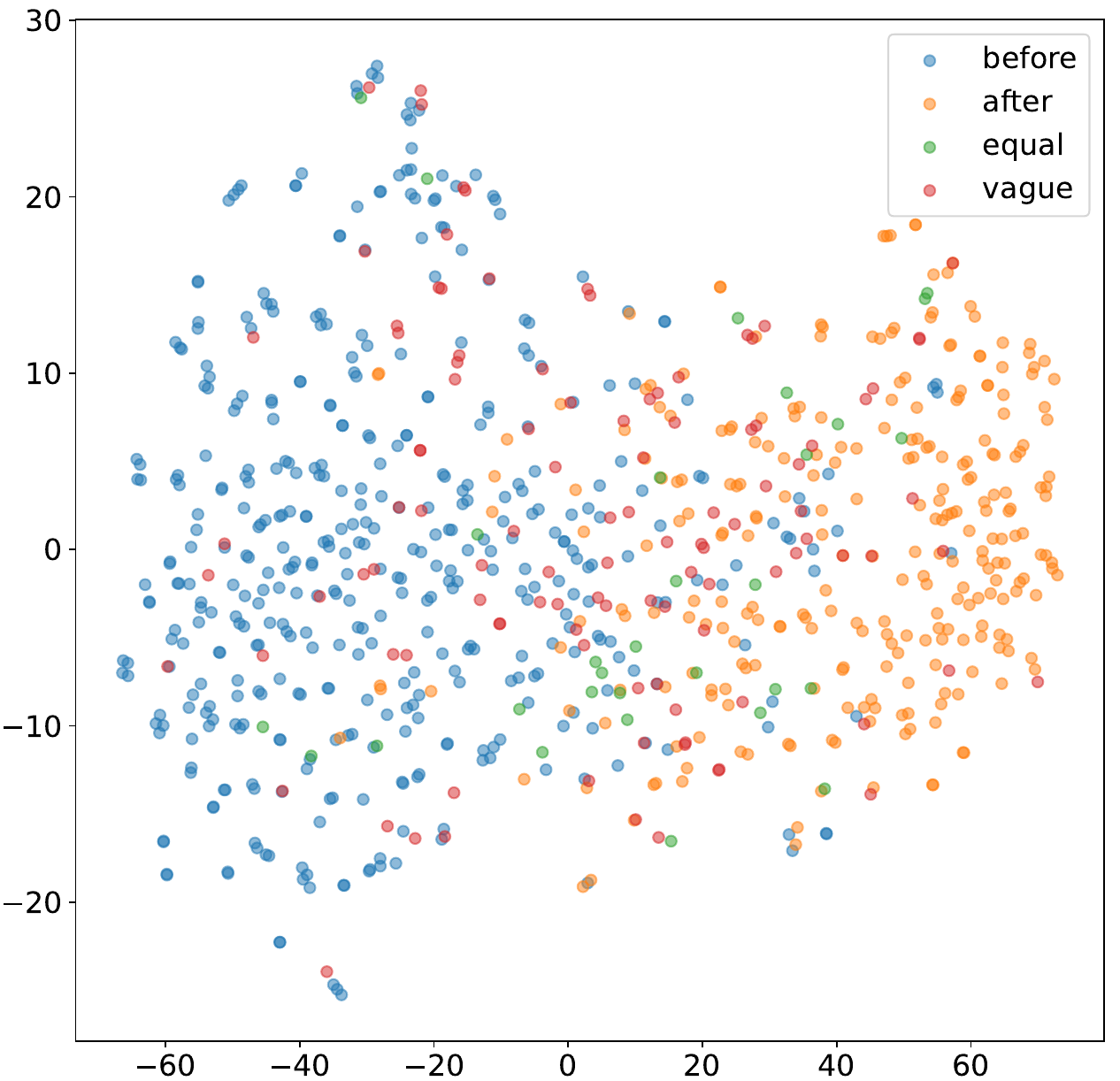}
\end{minipage}%
}%
\subfigure[After]{
\begin{minipage}[t]{0.48\linewidth}
\centering
\includegraphics[width=1.7in]{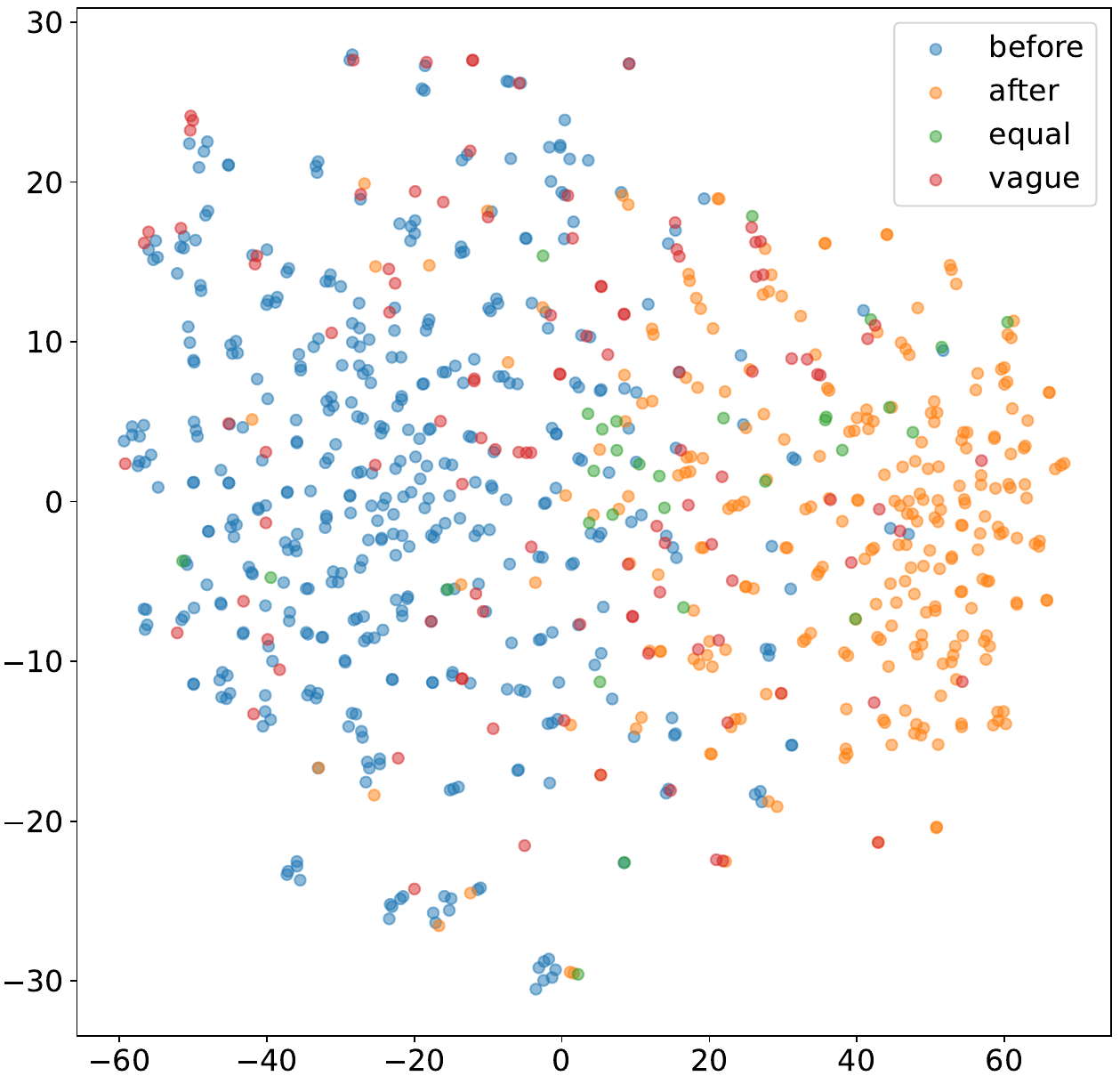}
\end{minipage}%
}%
\centering
\caption{Visualization of predictive representations before and after the introduction of contrastive learning. Each color indicates a temporal relation.}
\label{Visualization}
\end{figure}

\subsection{Visualization of Predicted Representations}
For the visual observation of the alleviating effect of contrastive learning on uneven data distribution, t-SNE \footnote{https://scikit-learn.org/stable/modules/generated/ sklearn.manifold.TSNE.html} is utilized to visualize the hidden vectors $\mathbf{h}_{[MASK]}$ at $[MASK]$ positions on the MATRES dataset. Fig. \ref{Visualization} illustrates the visualization results of the hidden vectors before and after the incorporation of contrastive learning. The inclusion of contrastive learning leads to a sharper classification boundary between \textit{BEFORE} and \textit{AFTER}. Additionally, the representations that belong to \textit{EQUAL} class are more densely clustered towards the midline between \textit{BEFORE} and \textit{AFTER}. However, the distribution of representations for the \textit{VAGUE} class is intricate and remains difficult to distinguish. As a result, contrastive learning facilitates the differentiation of classes with limited samples, yet there are opportunities for further investigation.

\subsection{Label-wise Analysis}
For a detailed analysis of the improvements made by our model in addressing data issues, we compare its performance with PSL on individual labels, as shown in Table \ref{label}. For labels with comparatively more samples, TemPrompt achieves improvements of 1.2\%, 11.7\% and 5.1\% in F1 scores for the \textit{VAGUE}, \textit{BEFORE} and \textit{AFTER} labels, respectively. In terms of labels with relatively fewer samples, despite a slight decrease for the \textit{INCLUDED IN} label, TemPrompt demonstrates an 8.3\% increase in F1 score compared to PSL for the \textit{INCLUDES} label. Especially, concerning the SIMULTANEOUS label, referencing recent works \cite{zhou2021clinical,zhang2022knowledge,ma2021eventplus}, TemPrompt achieves a remarkable advancement that raises the F1 scores from 0 to 34.1\%. Obviously, on the whole, our approach can effectively addresses the deficiencies and unevenness within human annotated data. However, it falls short of resolving all few-shot label challenges, highlighting the importance of further research.

\begin{table}[t]
  \centering
  \caption{F1 scores for each individual label on TB-Dense.}
  \label{label}
  \setlength{\tabcolsep}{3.5mm}{
  \resizebox{1\linewidth}{!}{
  \begin{tabular}{c|c|c}
    \toprule
    Labels& PSL & TemPrompt \\
    \midrule
    \textit{VAGUE} & 68.6 & \textbf{69.8}\\
    \textit{BEFORE} & 61.7 & \textbf{73.4}\\
    \textit{AFTER} & 70.7 & \textbf{75.8}\\
    \textit{INCLUDED IN} & \textbf{30.8} & 29.4\\
    \textit{INCLUDES} & 40.0 & \textbf{48.3} \\
    \textit{SIMULTANEOUS} & 0.0 & \textbf{34.1}\\
    \bottomrule
  \end{tabular}
}}
\end{table}

\subsection{Hyperparameter Analysis}
\begin{figure*}[htbp]
\centering 
\subfigure[$\tau$]{
\begin{minipage}[t]{0.32\linewidth}
\centering
\includegraphics[width=2.4in]{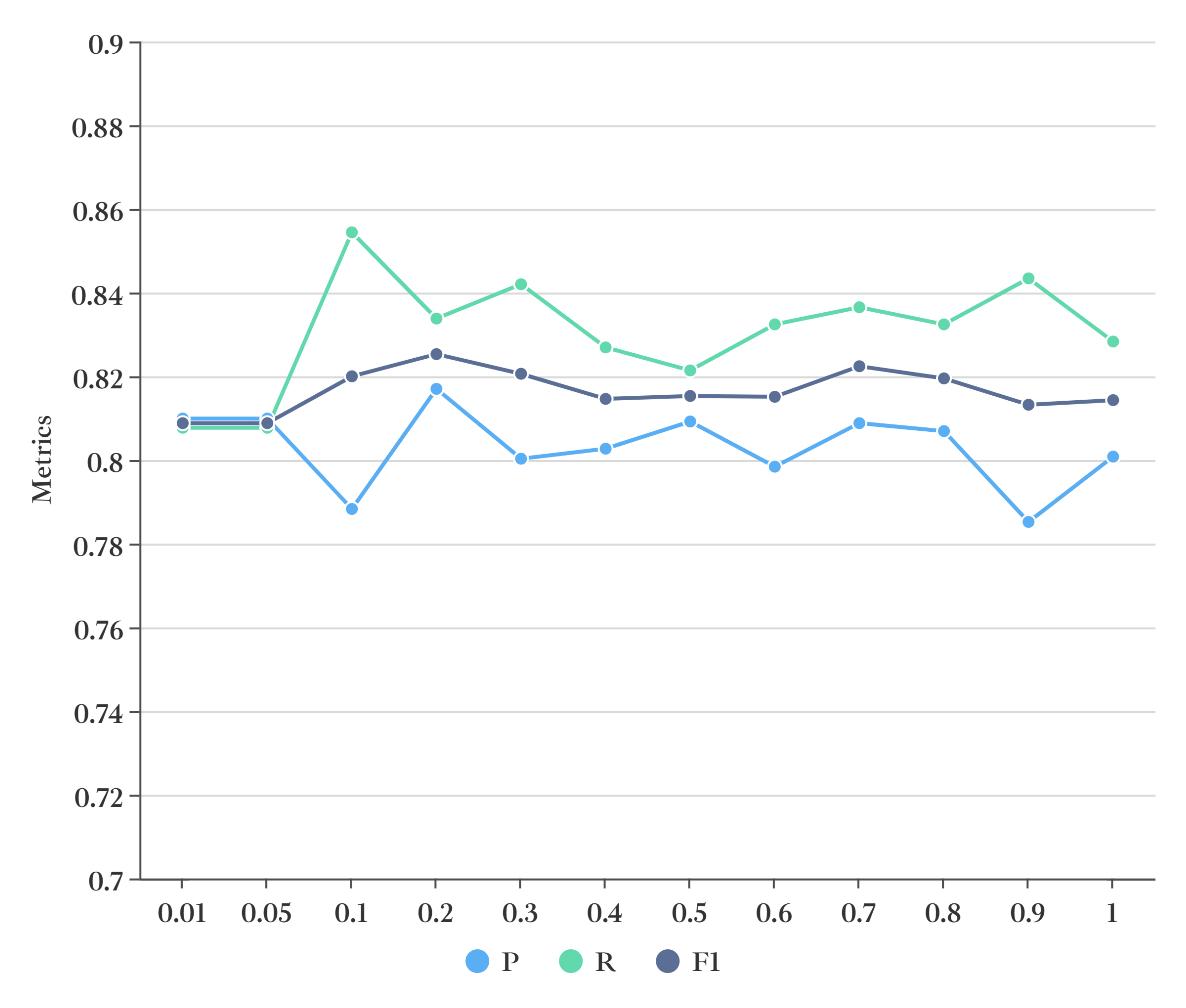}
\end{minipage}%
}%
\subfigure[$\alpha$]{
\begin{minipage}[t]{0.32\linewidth}
\centering
\includegraphics[width=2.4in]{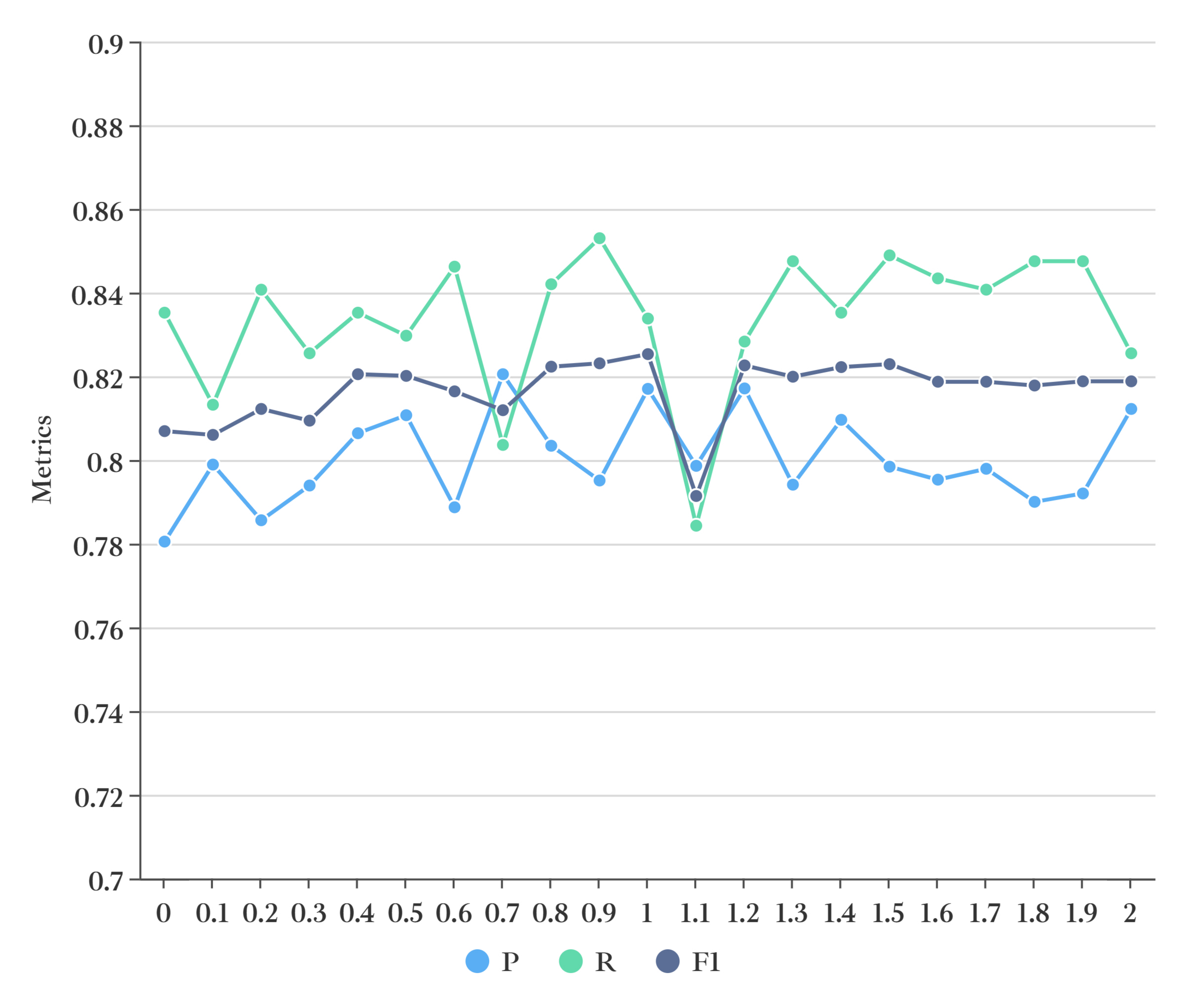}
\end{minipage}%
}%
\subfigure[$\beta$]{
\begin{minipage}[t]{0.32\linewidth}
\centering
\includegraphics[width=2.4in]{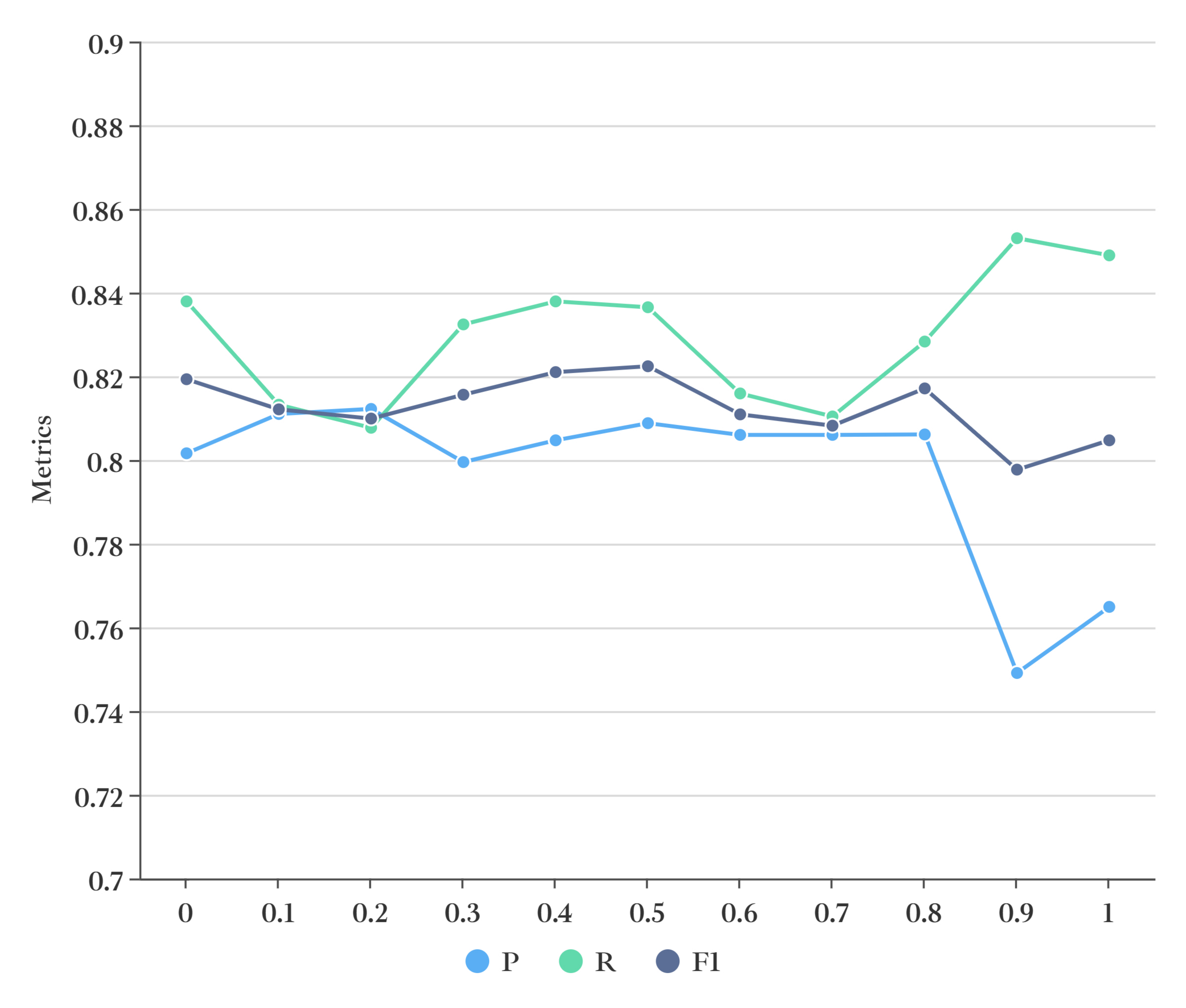}
\end{minipage}%
}%
\centering
\caption{Sensitivity analysis results of hyperparameters in P, R and F1. }
\label{para}
\end{figure*}

We investigate how different different values of hyperparameters $\tau$, $\alpha$ and $\beta$ affect the performance of TemPrompt using the MATRES dataset. Fig. \ref{para} depicts the variation in results as the hyperparameters change. In the general trend, the P-score and R-score exhibit a mirrored relation along a line parallel to the X-axis, where an increase in one leads to a decrease in the other. Therefore, we should seek a balance between these two scores to attain the optimal F1 score.

The temperature coefficient $\tau$ is utilized to regulate the model's discriminative ability towards negative samples in contrastive learning. Fig.\ref{para} (a) illustrates that the F1 score experiences two peaks as $\tau$ increases. If the hyperparameter $\tau$ is set too large, the similarity distribution becomes smoother, which treats all negative samples equally and leads to the model learning without emphasis. If $\tau$ is set too small, the model excessively focuses on extremely difficult negative samples, which could in fact be potential positive samples. This might hinder the model's convergence or result in limited generalization capability. Therefore, after careful consideration, we select $\tau=0.2$, corresponding to the optimal F1 score.

The hyperparameters $\alpha$ and $\beta$ aim to achieve a harmonious balance between the influences of $\mathcal{L}_{tre}$, $\mathcal{L}_{ter}$ and $\mathcal{L}_{con}$. As shown in Fig. \ref{para} (b) and (c), the F1 scores demonstrate a trend of initially increasing and then decreasing as both increase with the exception of $\alpha=1.1$. This is because excessively high hyperparameter values can obscure the focus of the task, while excessively low values make it difficult for the auxiliary task and contrastive learning to have an impact. Therefore, the hyperparameter values $\alpha=1$ and $\beta=0.5$ at the peaks are more appropriate.

\subsection{Case Study}
To explore the application of our approach in RAG-based crowdsourcing, we input sentences about printed circuit board (PCB) design and manufacturing, which is the outcome retrieved by the retriever in response to the queries posted by the requester, into the fine-tuned TemPrompt on TB-Dense. Note that the events mentioned in the sentence have been detected in advance. Table \ref{tab:example} presents the examples of TRE in the PCB manufacturing domain. Despite the lack of PCB-related data in the dataset, the model demonstrates its ability to effectively extract salient relations, i.e., \textit{BEFORE} and \textit{AFTER}. Nevertheless, the temporal relation between \textit{layout} and \textit{creation} is classified as \textit{BEFORE}, but conversely, it is erroneously categorized as \textit{SIMULTANEOUS}. This indicates one of the limitations of our model, prompting us to explore the integration of antisymmetric constraints in future research. For complex relations with subtle features, such as \textit{SIMULTANEOUS}, \textit{VAGUE}, \textit{INCLUDES} and \textit{INCLUDED IN}, we believe learning domain-specific knowledge is essential. For example, experts consider \textit{selection} and \textit{layout} to be in a continuous cycle of joint optimization, hence their temporal relation is annotated as \textit{SIMULTANEOUS}. Additionally, they view \textit{procurement} as an integral part of the \textit{manufacturing} process, so the relation between the former and the latter is \textit{INCLUDED IN} and conversely \textit{INCLUDES}. Obviously, these mistakes stem from a lack of expertise in the specific field. In our future work, we plan to collect domain-specific knowledge or data so that the retriever can access it and improve our model.

\begin{table}[]
  \centering
  \caption{Examples of TRE in specialized domains.}\label{tab:example}
  {
    \begin{tabularx}{1.0\linewidth}
    {  m{8.8cm}}
      \hline
      \textit{\textbf{Input}}: It begins with component \textit{(selection, e1)} and wiring \textit{(layout, e2)} using circuit design software, followed by the \textit{(creation, e3)} of a PCB schematic to represent the circuit connections.\\
       \textit{\textbf{Output}}:\\ r(e1,e2)=\small{\textit{VAGUE}} \, r(e1,e3)=\small{\textit{BEFORE}} \, r(e2,e3)=\small{\textit{BEFORE}}\\
      r(e2,e1)=\small{\textit{VAGUE}} \, r(e3,e1)=\small{\textit{AFTER}} \, r(e3,e2)=\small{\textit{SIMULTANEOUS}}\\
      \hline
         \textit{\textbf{Input}}: After \textit{(generating, e1)} Gerber files, the component \textit{(procurement, e2)} and PCB \textit{(manufacturing, e3)} processes are initiated.\\
         \textit{\textbf{Output}}: \\ r(e1,e2)=\small{\textit{BEFORE}} \,  r(e1,e3)=\small{\textit{BEFORE}} \,  r(e2,e3)=\small{\textit{SIMULTANEOUS}}\\
      r(e2,e1)=\small{\textit{AFTER}} \,  r(e3,e1)=\small{\textit{AFTER}}\,  r(e3,e2)=\small{\textit{SIMULTANEOUS}}\\
      \hline
    \end{tabularx}
  }
\end{table}

\section{Conclusion and Future Work}
In this paper, we propose TemPrompt, a novel framework that integrates prompt tuning and contrastive learning to address the scarcity and non-uniform distribution of annotated data for TRE. To comprehensively account for various factors pertinent to the TRE task, a task-oriented prompt construction approach is employed to automatically generate discrete cloze prompts. Additionally, inspired by the complementary between tasks, temporal event reasoning is introduced as an auxiliary task to deepen the model's grasp of temporal event knowledge. Extensive experiments on two datasets demonstrate that our method excels over current state-of-the-art methods across most metrics.  In the future, we plan to further explore the utilization of asymmetric constraints and domain-specific knowledge for the enhancement of our TRE model. Additionally, from the perspective of RAG, this paper is dedicated to designing a generator’s structures and assumes that the relevant task execution descriptions have already been obtained. However, the design of retrievers for this task remains unexplored, and this will be a focus of our future research.

\section*{CRediT authorship contribution statement}
\textbf{Jing Yang}: Conceptualization, Methodology, Writing - Original Draft, Writing - Review \& Editing. \textbf{Yu Zhao}: Methodology, Investigation, Validation. \textbf{Linyao Yang}: Writing - Review \& Editing. \textbf{Xiao Wang}: Writing - Review \& Editing, Funding acquisition. \textbf{Long Chen}: Writing - Review \& Editing. \textbf{Fei-Yue Wang}: Writing - Review \& Editing.

\section*{Declaration of competing interest}
The authors declare that they have no known competing financial interests or personal relationships that could have appeared to influence the work reported in this paper. 

\section*{Data availability}
Data will be made available on request. 

\section*{Acknowledgment}
This work was supported in part by the Zhejiang Lab Open Research Project under Grant K2022KG0AB02, and in part by the University Scientific Research Program of Anhui Province under Grant 2023AH020005.

\bibliographystyle{IEEEtran}
\bibliography{IFRef}

\end{document}